\documentclass{article}

\usepackage{microtype}
\usepackage{graphicx}
\usepackage{subfigure}
\usepackage{booktabs} % for professional tables

\usepackage{xr-hyper}
\usepackage{hyperref}

\usepackage{amssymb,amsmath,amstext,amstext,amsthm}
\usepackage{bm}
\usepackage{bbm}
\usepackage{makecell}
\usepackage{color}
\usepackage{wrapfig}
\usepackage[utf8]{inputenc} % allow utf-8 input
\usepackage[T1]{fontenc}    % use 8-bit T1 fonts

\usepackage{url}            % simple URL typesetting
\usepackage{booktabs}       % professional-quality tables
\usepackage{amsfonts}       % blackboard math symbols
\usepackage{nicefrac}       % compact symbols for 1/2, etc.
\usepackage{microtype}      % microtypography
\usepackage{xcolor}         % colors
\usepackage{enumitem}
\usepackage{algorithmic}
\usepackage{algorithm}
\usepackage{caption}
\usepackage{verbatim}
\newtheorem{theorem}{Theorem}
\newtheorem{lemma}{Lemma}
\newtheorem{proposition}{Proposition}

\usepackage[accepted]{icml2022}

\icmltitlerunning{Divergence-Regularized Multi-Agent Actor-Critic}

\begin{document}

\twocolumn[
\icmltitle{Divergence-Regularized Multi-Agent Actor-Critic}



\icmlsetsymbol{equal}{*}

\begin{icmlauthorlist}
\icmlauthor{Kefan Su}{to}
\icmlauthor{Zongqing Lu}{to}
\end{icmlauthorlist}

\icmlaffiliation{to}{School of Computer Science, Peking University}

\icmlcorrespondingauthor{Zongqing Lu}{zongqing.lu@pku.edu.cn}

\icmlkeywords{Machine Learning, ICML}

\vskip 0.3in
]



\printAffiliationsAndNotice{} % otherwise use the standard text.

\begin{abstract}
	Entropy regularization is a popular method in reinforcement learning (RL). Although it has many advantages, it alters the RL objective of the original Markov Decision Process (MDP). Though divergence regularization has been proposed to settle this problem, it cannot be trivially applied to cooperative multi-agent reinforcement learning (MARL). In this paper, we investigate divergence regularization in cooperative MARL and propose a \textit{novel} off-policy cooperative MARL framework, divergence-regularized multi-agent actor-critic (DMAC). Theoretically, we derive the update rule of DMAC which is naturally off-policy and guarantees monotonic policy improvement and convergence in both the original MDP and divergence-regularized MDP. We also give a bound of the discrepancy between the converged policy and optimal policy in the original MDP. DMAC is a flexible framework and can be combined with many existing MARL algorithms. Empirically, we evaluate DMAC in a didactic stochastic game and StarCraft Multi-Agent Challenge and show that DMAC substantially improves the performance of existing MARL algorithms. 

\end{abstract}

\section{Introduction}
\label{intro}

%Regularization is a common method for single-agent reinforcement learning (RL). The optimal policy learned by traditional RL algorithm is always deterministic \citep{RLBOOK}. This property may result in the inflexibility of the policy facing with unknown environments \citep{regularizer}. Entropy regularization is proposed to settle this problem by learning a policy according to the maximum-entropy principle \citep{deep_energy}. Moreover, entropy regularization is beneficial to exploration and robustness for RL algorithms \citep{SAC}. However, entropy regularization is imperfect. \citet{whatisquestion} pointed out maximum-entropy RL is a modification of the original RL objective because of the entropy regularizer. Maximum-entropy RL is actually learning an optimal policy for the entropy-regularized Markov Decision Process (MDP) rather than the original MDP, \textit{i.e.}, the converged policy may be biased. \citet{SAC-theory} analysed a more general case for regularization in RL and proposed what we call \textit{divergence regularization}. Divergence regularization can avoid the bias of the converged policy as well as be beneficial to exploration. \citet{DIV-AUG} employed divergence regularizer and proposed a single-agent RL algorithm, DAPO, which prevents the altering-objective drawback of entropy regularization.

Regularization is a common method for single-agent reinforcement learning (RL). The optimal policy learned by traditional RL algorithms is always deterministic \citep{RLBOOK}. This property may result in the inflexibility of the policy facing unknown environments \citep{regularizer}. Entropy regularization is proposed to settle this problem by learning a policy according to the maximum-entropy principle \citep{deep_energy}. Moreover, entropy regularization is beneficial to exploration and robustness for RL algorithms \citep{SAC}. However, entropy regularization is imperfect. \citet{whatisquestion} pointed out that maximum-entropy RL modifies the original RL objective because of the entropy regularizer. Maximum-entropy RL is actually learning an optimal policy for the entropy-regularized Markov Decision Process (MDP) rather than the original MDP, \textit{i.e.}, the converged policy may be biased. \citet{SAC-theory} analyzed a more general case for regularization in RL and proposed what we call \textit{divergence regularization}. Divergence regularization is beneficial to exploration and may be helpful to the bias issue. 

Regularization can also be applied to cooperative multi-agent reinforcement learning (MARL) \citep{transfer,FOP}. However, most cooperative MARL algorithms do not use regularizers \citep{MADDPG,COMA,QMIX,QTRAN,DGN,QPLEX}. Only few cooperative MARL algorithms such as FOP \citep{FOP} use entropy regularization, which may suffer from the drawback aforementioned. Divergence regularization, on the other hand, could potentially benefit cooperative MARL. In addition to its advantages mentioned above, divergence regularization can also help to control the step size of policy update which is similar to conservative policy iteration \citep{CPI} in single-agent RL. Conservative policy iteration and its successive methods such as TRPO \citep{TRPO} and PPO \citep{PPO} can stabilize policy improvement \citep{stable-policy}. These methods use a surrogate objective for policy update, but the policies in centralized training with decentralized execution (CTDE) paradigm may not preserve the properties of the surrogate objective. Moreover, DAPO \citep{DIV-AUG}, a single-agent RL algorithm using divergence regularizer,  cannot be trivially extended to cooperative MARL settings. Even with some tricks like V-trace \citep{V-trace} for off-policy correction, DAPO is essentially an on-policy algorithm and thus may not be sample-efficient in cooperative MARL settings.

In the paper, we propose divergence-regularized multi-agent actor-critic (DMAC), a \textit{novel} off-policy cooperative MARL framework. We analyze the general iteration of DMAC and theoretically show that DMAC guarantees the monotonic policy improvement and convergence in both the original MDP and the divergence-regularized MDP. We also derive a bound of the discrepancy between the converged policy and the optimal policy in the original MDP. Besides, DMAC is beneficial to exploration and stable policy improvement by applying our update rule of target policy. We also propose and analyze \textit{divergence policy iteration} in general cooperative MARL settings and a special case combined with value decomposition. Based on divergence policy iteration, we derive the off-policy update rule for the critic, policy, and target policy.  Moreover, DMAC is a flexible framework and can be combined with many existing cooperative MARL algorithms to substantially improve their performance.

%We empirically investigate DMAC in StarCraft Multi-Agent Challenge \citep{SMAC}. We combine DMAC with four representative MARL methods, \textit{i.e.}, COMA \citep{COMA} for on-policy multi-agent policy gradients, MAAC \citep{MAAC} for off-policy multi-agent actor-critic, QMIX \citep{QMIX} for value decomposition, and DOP \citep{DOP} for the combination of value decomposition and policy gradient. Experimental results show that DMAC indeed induces better performance, faster convergence, and better stability in most tasks, which verifies the benefits of DMAC and demonstrates the advantages of divergence regularizer over entropy regularizer in cooperative MARL.

We empirically investigate DMAC in a didactic stochastic game and StarCraft Multi-Agent Challenge \citep{SMAC}. We combine DMAC with five representative MARL methods, \textit{i.e.}, COMA \citep{COMA} for on-policy multi-agent policy gradient, MAAC \citep{MAAC} for off-policy multi-agent actor-critic, QMIX \citep{QMIX} for value decomposition, DOP \citep{DOP} for the combination of value decomposition and policy gradient, and FOP \citep{FOP} for the combination of value decomposition and entropy regularization. Experimental results show that DMAC indeed induces better performance, faster convergence, and better stability in most tasks, which verifies the benefits of DMAC and demonstrates the advantages of divergence regularization over entropy regularization in cooperative MARL.

\section{Related Work}

%\vspace{-0.2cm}
\textbf{MARL.} MARL has been a hot topic in the field of RL. In this paper, we focus on cooperative MARL. Cooperative MARL is usually modeled as Dec-POMDP \citep{Dec-POMDP}, where all agents share a reward and aim to maximize the long-term return. Centralized training with decentralized execution (CTDE) \citep{MADDPG} paradigm is widely used in cooperative MARL. CTDE usually utilizes a centralized value function to address the non-stationarity for multi-agent settings and decentralized policies for scalability. Many MARL algorithms adopt CTDE paradigm such as COMA, MAAC, QMIX, DOP, and FOP. COMA \citep{COMA} employs the counterfactual baseline which can reduce the variance as well as settle the credit assignment problem. MAAC \citep{MAAC} uses self-attention mechanism to integrate local observation and action of each agent and provides structured information for the centralized critic. Value decomposition \citep{VDN,QMIX,QTRAN,Qatten,QPLEX,DOP,FOP,DFAC,REQMIX,FACMAC} is a popular class of cooperative MARL algorithms. These methods express the global Q-function as a function of individual Q-functions to satisfy Individual-Global-Max (IGM), which means the optimal actions of individual Q-functions are corresponding to the optimal joint action of the global Q-function. QMIX \citep{QMIX} is a representative of value decomposition methods. It uses a hypernet to ensure the monotonicity of the global Q-function in terms of individual Q-functions, which is a sufficient condition of IGM. DOP \citep{DOP} is a method that combines value decomposition with policy gradient. DOP uses a linear value decomposition which is another sufficient condition of IGM and the linear value decomposition helps the compute of policy gradient. FOP \citep{FOP} is a method that combines value decomposition with entropy regularization and uses a more general condition, Individual-Global-Optimal, to replace IGM. In this paper, we will combine DMAC with these algorithms and show its improvement. MAPPO \citep{MAPPO} is a CTDE version of PPO \citep{PPO}, where a centralized state-value function is learned. However, the update rule of policy of MAPPO contradicts DMAC, so DMAC cannot be combined with MAPPO. 

%\textcolor{red}{MAPPO \citep{MAPPO} is a CTDE method which obtains a surprising effective and performance. However, MAPPO is actually more like a DTDE method with the centralized state information and has little theoretical guarantees. Moreover, the update rule of the policy of MAPPO contradicts with DMAC, so we don't take MAPPO as a baseline.}

\textbf{Regularization.} Entropy regularization was first proposed in single-agent RL. \citet{SAC-theory} analyzed the entropy-regularized MDP and revealed the properties about the optimal policy and the corresponding Q-function and V-function. They also showed the equivalence of value-based methods and policy-based methods in entropy-regularized MDP. \citet{SAC} pointed out maximum-entropy RL can achieve better exploration and stability facing the model and estimation error. Although entropy regularization has many advantages, \citet{whatisquestion} showed entropy regularization modifies the MDP and results in the bias of the convergent policy. \citet{regularizer} revealed the drawbacks of the convergent policy of general RL and maximum-entropy RL. The former is usually a deterministic policy \citep{RLBOOK} that is not flexible enough for unknown situations, while the latter is a policy with non-zero probability for all actions which may be dangerous in some scenarios. \citet{unifiedview} analyzed the entropy regularization method from several views. They revealed a more general form of regularization which is actually divergence regularization and showed entropy regularization is just a special case of divergence regularization. \citet{DIV-AUG} absorbed previous results and proposed an on-policy algorithm, \textit{i.e.}, DAPO. However, DAPO cannot be trivially applied to MARL. Moreover, its on-policy learning is not sample-efficient for MARL settings, and its off-policy correction trick V-trace \citep{V-trace} is also intractable in MARL. There are some previous studies in single-agent RL which use similar target policy to ours, but their purposes are quite different. Trust-PCL \citep{Trust-PCL} introduces a target policy as a trust-region constraint for maximum-entropy RL, but the policy is still biased by entropy regularizer. MIRL \citep{MIRL} uses a distribution that is only related to actions as the target policy to compute a mutual-information regularizer, but it still changes the objective of the original RL. 

%\vspace{-0.2cm}

\section{Preliminaries}
%\vspace{-0.2cm}
%\subsection{Dec-POMDP and Entropy Regularization}

%Markov Decision Process (MDP) is a basic tool for modeling in single-agent RL and used to describe the interaction between agent and environment. An MDP is a tuple $M=\left\{S,A,P,r,\gamma\right\}$, in which $s \in S$ is state, $a \in A$ is action, $P(s^{\prime} |s,a ): S \times A \times S \to [0,1]$ is transition probability from state $s$ to state $s^{\prime}$ after taking action $a$, $r(s,a):S \times A \to \mathbb{R}$ is the reward function, and $\gamma$ is the discount factor. Given an MDP, we can define a policy $\pi(a|s): S \times A \to [0,1]$, which is the probability of the agent takes action $a$ in state $s$.
\textbf{Dec-POMDP} is a general model for cooperative MARL. A Dec-POMDP is a tuple $M=\left\{S,A,P,Y,O,I,n,r,\gamma\right\}$. $S$ is the state space, $n$ is the number of agents, $\gamma$ is the discount factor, and $I = \{1,2\cdots n\}$ is the set of all agents.  $A = A_1 \times A_2 \times \cdots \times A_n$ represents the joint action space where $A_i$ is the individual action space for agent $i$. $P(s^{\prime} |s,\bm{a} ): S \times A \times S \to [0,1]$ is the transition function, and $r(s,\bm{a} ): S \times A \to \mathbb{R}$ is the reward function of state $s$ and joint action $\bm{a}$. $Y$ is the observation space, and $O(s,i):S \times I \to Y $ is a mapping from state to observation for each agent. The objective of Dec-POMDP is to maximize $J({\bm{\pi}}) = \mathbb{E}_{\bm{\pi}}\left[ \sum_{t = 0} \gamma^t r(s_t,\bm{a}_t ) \right],$ and thus we need to find the optimal joint policy ${\bm{\pi}}^{*} = \arg\max_{{\bm{\pi}}} J({\bm{\pi}})$. To settle the partial observable problem, history $\tau_i \in \mathcal{T}_i = (Y \times A_i)^*$ is often used to replace observation $o_i \in Y$. As for policies in CTDE, each agent $i$ has an individual policy $\pi_i(a_i|\tau_i)$ and the joint policy $\bm{\pi}$ is the product of each $\pi_i$. Though we calculate individual policy as $\pi_i(a_i|\tau_i)$ in practice, we will use $\pi_i(a_i|s)$ in analysis and proofs for simplicity.
%\footnote{$X^*$ means the set of all possible sequences generated by the elements of the set $X$.}

\textbf{Entropy regularization} adds the logarithm of the probability of the sampled action according to current policy to the reward function. It modifies the optimization objective as 
$J_{\operatorname{ent}}(\bm{\pi}) = \mathbb{E}_{\bm{\pi}}\left[ \sum_{t = 0} \gamma^t \left( r(s_t,\bm{a}_t ) -\omega \log \bm{\pi}(\bm{a}_t|s_t) \right)  \right].$
We also have the corresponding Q-function $Q_{\operatorname{ent}}^{\bm{\pi}}(s,\bm{a}) = r(s,\bm{a}) + \gamma \mathbb{E}\left[ V_{\operatorname{ent}}^{\bm{\pi}}(s^\prime) \right]$ and V-function $V_{\operatorname{ent}}^{\bm{\pi}}(s)= \mathbb{E}_{\bm{\pi}}\left[ \sum_{t = 0} \gamma^t \left( r(s_t,\bm{a}_t ) -\omega \log \bm{\pi}(\bm{a}_t|s_t) \right) | s_0 = s \right]$.
Given these definitions, we can deduce an interesting property $V_{\operatorname{ent}}^{\bm{\pi}}(s) = \mathbb{E}\left[Q_{\operatorname{ent}}^{\bm{\pi}}(s,\bm{a})\right] + \omega \mathcal{H}\left( \bm{\pi}(\cdot|s)\right)$, where $\mathcal{H}\left( \bm{\pi}(\cdot|s)\right)$ represents the entropy of policy $\bm{\pi}(\cdot|s)$. $V_{\operatorname{ent}}^{\bm{\pi}}(s)$ includes an entropy term which is the reason it is called \textit{entropy regularization}.

%Dec-POMDP is a general model for cooperative MARL. A Dec-POMDP is a tuple $M=\left\{S,A,P,Y,O,I,n,r,\gamma\right\}$, where $S,P,r,\gamma$ are similar as in MDP. $A = A_1 \times A_2 \times \cdots \times A_n$ represents the joint action space where $A_i$ is the individual action space for agent $i$, $P(s^{\prime} |s,\bm{a} ): S \times A \times S \to [0,1]$ is the transition function, $r(s,\bm{a} ): S \times A \to \mathbb{R}$ is the reward function of joint action $\bm{a}$, $n$ is the number of agents, and $I$ is the set of all agents.

%In practice, we usually take an approximation to calculate individual policy, \textit{i.e.}, $\pi_i(a_i|\tau_i) \approx \pi_i(a_i|s)$. 

%\vspace{-0.2cm}

\section{Method}
In this section, we first give the definition of divergence regularization. Then we propose the selection of target policy and derive its theoretical properties. Next we propose and analyze \textit{divergence policy iteration}. Finally, we derive the update rules of the critic and actors and obtain the algorithm of DMAC. Note that all proofs are given in Appendix~\ref{app:proofs}.

%In this section, we first give the definition of divergence regularization. Then we introduce our method in general and discuss the advantage of the divergence regularization partly. Next we propose and analyze \textit{divergence policy iteration}. Finally, based on divergence policy iteration, we derive the update rules of the critic, policy, and target policy for divergence-regularized MARL and obtain the algorithm of DMAC.

\subsection{Divergence Regularization}
%DMAC adds a target policy to obtain the divergence-regularized MDP. 
We maintain a target policy $\rho_i$ for each agent $i$, which is different from the agent policy $\pi_i$. Together we have a joint target policy ${\bm{\rho}} = \prod_{i=1}^n \rho_i$. This joint target policy ${\bm{\rho}}$ modifies the objective function as 
$J_{\bm{\rho}}({\bm{\pi}}) = \mathbb{E}_{\bm{\pi}}\left[ \sum_{t = 0} \gamma^t \left(r(s_t,\bm{a}_t ) - \omega \log \frac{{\bm{\pi}}(\bm{a}_t|s_t)}{{\bm{\rho}}(\bm{a}_t|s_t)} \right)  \right].$
That is, a regularizer $\log\frac{{\bm{\pi}}(\bm{a}_t|s_t)}{{\bm{\rho}}(\bm{a}_t|s_t)}$, which describes the \textit{discrepancy} between policy ${\bm{\pi}}$ and target policy ${\bm{\rho}}$, is added to the reward function just like entropy regularization.

Given $\bm{\rho}$, we can define the corresponding V-function and Q-function for divergence regularization as follows,
\begin{align*}
	& V_{\bm{\rho}}^{{\bm{\pi}}}(s) = \mathbb{E}_{\bm{\pi}}\left[ \sum_{t = 0} \gamma^t (r(s_t,\bm{a}_t ) - \omega \log \frac{{\bm{\pi}}(\bm{a}_t|s_t)}{{\bm{\rho}}(\bm{a}_t|s_t)} ) | s_0 = s \right] \\
	& Q_{\bm{\rho}}^{{\bm{\pi}}}(s,\bm{a}) = r(s,\bm{a}) + \gamma \mathbb{E}_{s^\prime \sim P(\cdot | s, a)}\left[ V_{\bm{\rho}}^{\bm{\pi}}(s^\prime) \right].
\end{align*} 

Further, by simple deduction, we have
\begin{align*}
	V_{\bm{\rho}}^{\bm{\pi}}(s) 
	& = \mathbb{E}_{\bm{a} \sim \bm{\pi}(\cdot| s)}\left[ Q_{\bm{\rho}}^{{\bm{\pi}}}(s,\bm{a}) - \omega \log \frac{{\bm{\pi}}(\bm{a}|s)}{{\bm{\rho}}(\bm{a}|s)}\right] \\
	& = \mathbb{E}_{\bm{a} \sim \bm{\pi}(\cdot| s)}\left[Q_{\bm{\rho}}^{\bm{\pi}}(s,\bm{a}) \right] - \omega D_{\operatorname{KL}}\left( {\bm{\pi}}(\cdot|s) \| {\bm{\rho}}(\cdot|s) \right). \nonumber
\end{align*}
$V_{\bm{\rho}}^{\bm{\pi}}(s)$ includes an extra term which is the KL divergence between ${\bm{\pi}}$ and ${\bm{\rho}}$, and thus this regularizer is referred to as \textit{divergence regularization}.

\subsection{Target Policy}

%We have discussed the update rule of the critic and actors, and now we focus on the selection and update rule of the target policy. With the update rules above, we can obtain divergence policy iteration given a fixed target policy $\bm{\rho}$. Then we need to devise the update rule of $\bm{\rho}$ to prevent the bias of regularization and benefit the learning procedure.

%\textcolor{red}{We will first introduce our general method and our choice of the target policy in this method. Then we will discuss the benefit of the target policy and show some theoretical results about our method. }

Intuitively, this regularizer $\log\frac{{\bm{\pi}}(\bm{a}_t|s_t)}{{\bm{\rho}}(\bm{a}_t|s_t)}$ could help to balance exploration and exploitation. For example, for some action $\bm{a}$, if ${\bm{\rho}}(\bm{a}|s) > {\bm{\pi}}(\bm{a}|s)$, then the regularizer is equivalent to adding a positive value to the reward and \textit{vice versa}. Therefore, if we choose previous policy as target policy, the regularizer will encourage agents to take actions whose probability has decreased and discourage agents to take actions whose probability has increased. Additionally, the regularizer actually controls the discrepancy between current policy and previous policy, which could stabilize the policy improvement \citep{CPI,TRPO,PPO}.

We further analyze the selection of target policy theoretically. Let $\mu_{\bm{\pi}}$ denote the state-action distribution given a policy $\bm{\pi}$. That is, $\mu_{\bm{\pi}}(s,\bm{a}) = d^{\bm{\pi}}(s) {\bm{\pi}}(\bm{a}|s )$, where $d^{\bm{\pi}}(s) = \sum_{t = 0} \gamma^t \operatorname{Pr}(s_t = s| {\bm{\pi}})$ is the stationary distribution of states given ${\bm{\pi}}$. 
%Thus, we also have ${\bm{\pi}}_\mu(\bm{a}|s) = \mu(s,\bm{a})/\sum_{\bm{b} } \mu(s,\bm{b})$.
With $\mu_{\bm{\pi}}$, we can rewrite the optimization objective $J_{\bm{\rho}}({\bm{\pi}})$ as follows,
\begin{align}
	J_{\bm{\rho}}({\bm{\pi}}) & = \sum_{s,\bm{a}} \mu_{\bm{\pi}}(s,\bm{a}) r(s,\bm{a}) - \omega \sum_{s,\bm{a}} \mu_{\bm{\pi}}(s,\bm{a}) \log \frac{{\bm{\pi}}(\bm{a}|s)}{{\bm{\rho}}(\bm{a}|s)} \nonumber \\
	& = \sum_{s,\bm{a}} \mu_{\bm{\pi}}(s,\bm{a}) r(s,\bm{a}) - \omega D_{\operatorname{C}} \left( \mu_{\bm{\pi}} \| \mu_{\bm{\rho}} \right), \label{alter-J}
\end{align} 
where $D_{\operatorname{C}} \left( \mu_{\bm{\pi}} \| \mu_{\bm{\rho}} \right) = \sum_{s,\bm{a}} \mu_{\bm{\pi}}(s,\bm{a}) \log \frac{{\bm{\pi}}(\bm{a}|s)}{{\bm{\rho}}(\bm{a}|s)}$ is a Bregman divergence \citep{unifiedview}. Therefore, the objective of the divergence-regularized MDP can be expressed as 
\begin{equation} \label{MD-process}
	{\bm{\pi}}_{\bm{\rho}}^* = \arg \max_{{\bm{\pi}}} \sum_{s,\bm{a}} \mu_{\bm{\pi}}(s,\bm{a}) r(s,\bm{a}) - \omega D_{\operatorname{C}} \left( \mu_{\bm{\pi}} \| \mu_{\bm{\rho}} \right).
	%\vspace{-0.2cm}
\end{equation}
With this property, similar to \citet{unifiedview} and \citet{DIV-AUG}, we can use the following iterative process,
\begin{equation}
	\label{mirror}
	\bm{\pi}^{t+1} = \arg \max_{{\bm{\pi}}} \sum_{s,\bm{a}} \mu_{\bm{\pi}}(s,\bm{a}) r(s,\bm{a})  - \omega D_{\operatorname{C}} \left( \mu_{\bm{\pi}} \| \mu_{{\bm{\pi}}^t} \right). 
\end{equation}
% \begin{equation}\label{mirror}
%     \resizebox{1\hsize}{!}{%
%     $\bm{\pi}^{t+1} = \arg \underset{\bm{\pi}}{\max} \underset{s,\bm{a}}{\sum} \mu_{\bm{\pi}}(s,\bm{a}) r(s,\bm{a})  - \omega D_{\operatorname{C}} \left( \mu_{\bm{\pi}} \| \mu_{{\bm{\pi}}^t} \right)$%
%     } 
% \end{equation}
%This iteration is a mirror descent process \citep{unifiedview}, so the convergence of the policy is guaranteed. This process also guarantees that when the policy converges, $D_{\operatorname{C}} \left( \mu_{{\bm{\pi}}^{t+1}} \| \mu_{{\bm{\pi}}^t} \right) \to 0$; \textit{i.e.}, the regularizer will vanish. Moreover, we can obtain the following inequalities:
In this iteration, by taking the target policy $\bm{\rho}$ as the previous policy $\bm{\pi}^t$ when updating the policy $\bm{\pi}^{t+1}$, we can obtain the following inequalities,
\iffalse
$$
J(\boldsymbol{\pi}^{t + 1}) \ge J(\boldsymbol{\pi}^{t + 1}) - \omega D_{\operatorname{C}} \left( \mu_{\boldsymbol{\pi}^{t+1}} \| \mu_{{\boldsymbol{\pi}}^t} \right) \ge J(\boldsymbol{\pi}^{t }) - \omega D_{\operatorname{C}} \left( \mu_{\boldsymbol{\pi}^{t}} \| \mu_{{\boldsymbol{\pi}}^t} \right) = J(\boldsymbol{\pi}^{t }).
$$
\fi
\begin{align*}
    J(\boldsymbol{\pi}^{t + 1})
    & \ge J(\boldsymbol{\pi}^{t + 1}) - \omega D_{\operatorname{C}} \left( \mu_{\boldsymbol{\pi}^{t+1}} \| \mu_{{\boldsymbol{\pi}}^t} \right) =  J_{{\boldsymbol{\pi}}^t}(\boldsymbol{\pi}^{t + 1}) \\ 
    & \ge J(\boldsymbol{\pi}^{t }) - \omega D_{\operatorname{C}} \left( \mu_{\boldsymbol{\pi}^{t}} \| \mu_{{\boldsymbol{\pi}}^t} \right) \\
    & =  J(\boldsymbol{\pi}^{t}) \ge  J_{{\boldsymbol{\pi}}^{t-1}}(\boldsymbol{\pi}^{t}).
%\vspace{-0.2cm}
\end{align*}
The first and the third inequalities are from $D_{\operatorname{C}} \left( \mu_{\boldsymbol{\pi}} \| \mu_{{\boldsymbol{\rho}}} \right) \ge 0$, and the second inequality is from the definition of $\boldsymbol{\pi}^{t + 1}$. This indicates the policy sequence obtained by this iteration improves monotonically in both the divergence-regularized MDP (\textit{i.e.}, $ J_{{\boldsymbol{\pi}}^t}(\boldsymbol{\pi}^{t + 1}) \ge J_{{\boldsymbol{\pi}}^{t-1}}(\boldsymbol{\pi}^{t})$) and the original MDP (\textit{i.e.}, $ J(\boldsymbol{\pi}^{t+1}) \ge J(\boldsymbol{\pi}^{t})$). Moreover, as the sequences $\{ J_{{\boldsymbol{\pi}}^t}(\boldsymbol{\pi}^{t + 1})\}$ and $\{ J(\boldsymbol{\pi}^{t + 1})\}$ are both bounded, these two sequences will converge. With these deductions, we can obtain the following proposition.
%\vspace{-0.2cm}
\begin{proposition} \label{monotonic_converge}
    By iteratively applying the iteration \eqref{mirror} and taking $\bm{\pi}^{k}$ as $\bm{\rho}^k$, the policy sequence $\{\bm{\pi}^k\}$ converges and improves monotonically in both the divergence-regularized MDP and the original MDP.
\end{proposition} 
For the sake of simplicity, we denote the converged policy of iteration \eqref{mirror} and its corresponding Q-function and V-function as $\tilde{\bm{\pi}}^{*}$, $\tilde{Q}^{*}$ and $\tilde{V}^{*}$, respectively. Then we further discuss the property of $\tilde{\bm{\pi}}^{*}$, $\tilde{Q}^{*}$ and $\tilde{V}^{*}$.
Actually, the expression of $\tilde{\bm{\pi}}^{*}$ is decided by the initial policy $\bm{\pi}^0$ and the action that obtains the optimal value of $\tilde{Q}^{*}$. We define the set containing the optimal actions as $U_s = \{\bm{a}^\prime \in A | \bm{a}^\prime = \arg\max_{\bm{a} } \tilde{Q}^{*}(s,\bm{a} )\}$. Without loss of generality, we suppose that $\bm{\pi}^0(\bm{a}|s)>0$ for all state-action pairs. Then we have the following proposition about $\tilde{\bm{\pi}}^{*}$.
\begin{proposition} \label{converged_policy}
        %\vspace{-0.4cm}
    %\begin{equation}
            %\vspace{-0.4cm}
All the probabilities of $\tilde{\bm{\pi}}^{*}$ lie in the optimal actions of $\tilde{Q}^{*}$ and are proportional to their probabilities in the initial policy $\bm{\pi}^0$,
        $$\tilde{\bm{\pi}}^{*}(\bm{a}|s) = \mathbbm{1}{(\bm{a} \in U_s)} \frac{\bm{\pi}^0(\bm{a}|s)}{\sum_{\bm{a}^\prime \in U_s}\bm{\pi}^0(\bm{a}^\prime|s)}.$$
    %\end{equation}
\end{proposition}
%\textcolor{red}{The proof is included in the Appendix \ref{app:converged_policy}. The proposition \ref{converged_policy} tells us that all the probabilities of $\tilde{\bm{\pi}}^{*}$ lie in the optimal actions of $\tilde{Q}^{*}$ and are proportional to the value of initial policy $\bm{\pi}^0$.}
%since the policy sequence $\{\bm{\pi}^t\}$ converges to $\tilde{\bm{\pi}}^{*}$, the regularization term $\log \frac{\bm{\pi}^{t+1}(\bm{a}|s)}{\bm{\pi}^t(\bm{a}|s)}$ converges to $0$.
As for the property of the $\tilde{Q}^{*}$ and $\tilde{V}^{*}$,  by the definition of $\tilde{Q}^{*}$ and $\tilde{V}^{*}$, we have the following proposition. 
\begin{proposition} \label{converged_Q}
$\tilde{Q}^*$ is the same as $Q^{\tilde{\bm{\pi}}^*}$, the Q-function of the policy $\tilde{\bm{\pi}}^{*}$ in the original MDP, while $\tilde{V}^*$ is the same as $V^{\tilde{\bm{\pi}}^* }$, the V-function of the policy $\tilde{\bm{\pi}}^{*}$ in the original MDP.

%and $\tilde{V}^*$ are respectively the same as the Q-function $Q^{\tilde{\bm{\pi}}^*}$ and V-function $V^{\tilde{\bm{\pi}}^* }$ of the policy $\tilde{\bm{\pi}}^{*}$ in the original MDP.
    %\vspace{-0.4cm}
    %\begin{equation*}
    %    \tilde{Q}^* = Q^{\tilde{\bm{\pi}}^* }, \tilde{V}^* = V^{\tilde{\bm{\pi}}^* }
        %\vspace{-0.4cm}
    %\end{equation*}
\end{proposition}
With all these results above, we can further derive the discrepancy between the policy $\tilde{\bm{\pi}}^{*}$ and the optimal policy $\bm{\pi}^*$ in the original MDP in terms of their V-functions, and have the following theorem.
%\vspace{-0.4cm}
\begin{theorem} \label{theorem:global_optimal}
    If the initial policy $\bm{\pi}^0$ is a uniform distribution, then we have 
            %\vspace{-0.4cm}
    \begin{equation}\label{eq:global_optimal}
            %\vspace{-0.2cm}
        \Vert V^{\tilde{\bm{\pi}}^*} - V^* \Vert_{\infty} \le \frac{\omega}{1 - \gamma} \log |A|, 
    \end{equation}
    where $V^*$ is the optimal V-function in the original MDP and $|A|$ is the number of actions in $A$.
\end{theorem}
This theorem tells us that the discrepancy between $\tilde{\bm{\pi}}^{*}$ and $\bm{\pi}^*$ can be controlled by the coefficient of the regularization term, $\omega$.

%The regularization term vanishes in the converged Q-function and V-function, so it will not bias the converged policy.
At this stage, we can partly conclude the benefits of the divergence regularization or the target policy. The divergence regularization is beneficial to exploration which can be witnessed from the empirical results later. The policy sequence obtained by our selection of target policy converges and monotonically improves \textit{not just in the regularized MDP, but also in the original MDP}. Moreover, the V-function of the converged policy in the \textit{original} MDP can be sufficiently approximate to that of the optimal policy with a proper $\omega$.
%\vspace{-0.2cm}

%DMAC can be easily combined with existing MARL algorithms which is shown in the experiment.

\subsection{Divergence Policy Iteration}

To complete the update in the iteration \eqref{mirror}, we need to study the divergence-regularized MDP, given a fixed target policy $\bm{\rho}$. 

From the perspective of policy evaluation, we can define an operator $\Gamma^{\operatorname{\bm{\pi}}}_{\operatorname{\bm{\rho}}}$ as
\iffalse
\begin{equation*}
	\Gamma^{\operatorname{\bm{\pi}}}_{\operatorname{\bm{\rho}}} Q(s,\bm{a}) = r(s,\bm{a}) + \gamma \mathbb{E}_{s^{\prime} \sim P(\cdot | s, \bm{a}), \bm{a^{\prime}} \sim  \bm{\pi}(\cdot | s^{\prime}) } \left[ Q(s^{\prime},\bm{a^{\prime}}) - \omega \log \frac{{\bm{\pi}}(\bm{a^{\prime}}|s^{\prime})}{{\bm{\rho}}(\bm{a^{\prime}}|s^{\prime})} \right]
\end{equation*}
\fi
\begin{equation*}
	\Gamma^{\operatorname{\bm{\pi}}}_{\operatorname{\bm{\rho}}} Q(s,\bm{a}) = r(s,\bm{a}) + \gamma \mathbb{E} \left[ Q(s^{\prime},\bm{a^{\prime}}) - \omega \log \frac{{\bm{\pi}}(\bm{a^{\prime}}|s^{\prime})}{{\bm{\rho}}(\bm{a^{\prime}}|s^{\prime})} \right]
\end{equation*}
and have the following lemma. 

\begin{lemma}[{\bf Divergence Policy Evaluation}] \label{policy_evaluation}
	For any initial Q-function $Q^{0}(s,\bm{a}): S \times A \to \mathbb{R} $, we define a sequence $\left\{ Q^{k} \right\}$ given operator $\Gamma^{\operatorname{\bm{\pi}}}_{\operatorname{\bm{\rho}}}$ as 
	$
	Q^{k +1} = \Gamma^{\operatorname{\bm{\pi}}}_{\operatorname{\bm{\rho}}} Q^{k}
	$.
	Then, the sequence will converge to $Q_{\bm{\rho}}^{\bm{\pi}}$ as $k \to \infty$.
\end{lemma}
%The proof of Theorem \ref{policy_evaluation} is given in Appendix \ref{sec:app-eva}.

After the evaluation of the policy, we need a way to improve the policy. We have the following lemma about policy improvement.

\begin{lemma}[{\bf Divergence Policy Improvement}] \label{policy-theorem}
	If we define ${\bm{\pi}}_{\operatorname{new}}$ satisfying
	\begin{equation}\label{eq:lemma2}
		{\bm{\pi}}_{\operatorname{new}}(\cdot|s) = \arg \min_{\bm{\pi}} D_{\operatorname{KL}}\left( {\bm{\pi}}(\cdot|s) \| \bm{u}(\cdot|s) \right),
	\end{equation}
	where $\bm{u}(\cdot|s) = {\bm{\rho}}(\cdot|s) \frac{\exp\left( Q^{{\bm{\pi}}_{\operatorname{old}}}_{\bm{\rho}}(s,\cdot) / \omega \right)}{Z^{{\bm{\pi}}_{\operatorname{old}}}(s)}$ and $Z^{{\bm{\pi}}_{\operatorname{old}}}(s)$ is a normalization term, then for all actions $\bm{a}$ and all states $s$ we have
	$Q^{{\bm{\pi}}_{\operatorname{new}}}_{\bm{\rho}}(s,\bm{a}) \geq Q^{{\bm{\pi}}_{\operatorname{old}}}_{\bm{\rho}}(s,\bm{a})$.
\end{lemma}

%The proof of Theorem \ref{policy-theorem} is given in Appendix \ref{sec:app-improv}. 
Lemma \ref{policy_evaluation} and \ref{policy-theorem} can be seen as corollaries of the conclusion of \citet{SAC}. Lemma \ref{policy-theorem} indicates that given a policy ${\bm{\pi}}_{\operatorname{old}}$, if we find a policy ${\bm{\pi}}_{\operatorname{new}}$ according to (\ref{eq:lemma2}), then the policy ${\bm{\pi}}_{\operatorname{new}}$ is better than ${\bm{\pi}}_{\operatorname{old}}$. 

Lemma \ref{policy-theorem} does not make any assumption and is for general settings. Further, the policy improvement can be established and simplified based on value decomposition. 	In the following, we give an example for linear value decomposition (LVD) like DOP (\textit{i.e.}, $Q(s,\bm{a}) = \sum_i k_i(s) Q_i(s,a_i) + b(s)$) \citep{DOP}. 

%Theorem \ref{policy-theorem} is actually for general situations. If we have more assumptions such as linear value decomposition like DOP \citep{DOP}, \textit{i.e.}, $Q(s,\bm{a}) = \sum_i k_i(s) Q_i(s,a_i) + b(s)$, then we will have a simpler formula as followings.

\begin{lemma}[\bf Divergence Policy Improvement with LVD] \label{policy-LVD-theorem}
	If Q-functions satisfy $Q^{{\bm{\pi}}}_{\bm{\rho}}(s,\bm{a}) = \sum_i k_i(s) Q^{\pi_i}_{\bm{\rho}}(s,a_i) + b(s)$ and  we define ${\pi}_{\operatorname{new}}^i$ satisfying
	\[\pi^i_{\operatorname{new}}(\cdot|s) = \arg \min_{\pi_i} D_{\operatorname{KL}}\left( \pi_i(\cdot|s) \| u_i(\cdot|s) \right) \quad \forall i \in I,\]
	where $u_i(\cdot|s) = {\rho_i}(\cdot|s) \frac{\exp\big( k_i(s) Q^{{\pi}^i_{\operatorname{old}}}_{\bm{\rho}}(s,\cdot) / \omega \big)}{Z^{{\pi}^i_{\operatorname{old}}}(s)}$ and $Z^{{\pi}^i_{\operatorname{old}}}(s)$ is a normalization term, then for all actions $\bm{a}$ and all states $s$, we have
	$Q^{{\bm{\pi}}_{\operatorname{new}}}_{\bm{\rho}}(s,\bm{a}) \geq Q^{{\bm{\pi}}_{\operatorname{old}}}_{\bm{\rho}}(s,\bm{a})$.
\end{lemma}

Lemma \ref{policy-LVD-theorem} further tells us that if the MARL setting satisfies the condition of linear value decomposition, then each agent can optimize its individual policy with an objective of its own individual Q-function, which immediately improves the joint policy. 
%The proof of Theorem \ref{policy-LVD-theorem} is included in Appendix \ref{sec:app-them-LVD}. 
By combining divergence policy evaluation and divergence policy improvement, we have the following theorem of \textit{divergence policy iteration}.

\begin{theorem}[{\bf Divergence Policy Iteration}] \label{policy_iteration}
	By iteratively using Divergence Policy Evaluation and Divergence Policy Improvement, we will get a sequence $\left\{ Q^{k} \right\}$ and this sequence will converge to the optimal Q-function $Q_{\bm{\rho}}^{*}$ and the corresponding policy sequence will converge to the optimal policy $\bm{\pi}^{*}_{\bm{\rho}}$.
\end{theorem}

%The proof of Theorem \ref{policy_iteration} is given in Appendix \ref{sec:app-iter}.
Theorem \ref{policy_iteration} shows that with repeated application of divergence policy improvement and divergence policy evaluation, the policy can \textit{monotonically} improve and converge to the optimal policy. We use ${\bm{\pi}}^*_{\bm{\rho}}$, $V_{\bm{\rho}}^*(s)$, and $Q_{\bm{\rho}}^{*}(s,\bm{a})$ to denote the optimal policy, Q-function, and V-function, respectively, given a target policy ${\bm{\rho}}$.

With all these results above, we have enough tools to obtain the practical update rule of the critic and actors of DMAC.

\subsection{Divergence-Regularized Critic}

For the update of the critic, we have the following proposition.
\begin{proposition} \label{Q_coro}
    \[Q^*_{\bm{\rho}}(s,\bm{a}) = r(s,\bm{a}) + \gamma \mathbb{E}\left[ Q^*_{\bm{\rho}}(s^{\prime},\bm{a}^{\prime}) - \omega \log \frac{{\bm{\pi}^*_{\bm{\rho}}}(\bm{a}^{\prime}|s^{\prime})}{{\bm{\rho}}(\bm{a}^{\prime}|s^{\prime})}  \right] \]
	is tenable for all actions $\bm{a}^{\prime} \in A$.
\end{proposition} 
\iffalse
\[Q^*_{\bm{\rho}}(s,\bm{a}) = r(s,\bm{a}) + \gamma \mathbb{E}_{s^\prime \sim P(\cdot|s, \bm{a}), \bm{a}^\prime \sim \bm{\pi}^*_{\bm{\rho}}(\cdot | s^\prime ) }\left[ Q^*_{\bm{\rho}}(s^{\prime},\bm{a}^{\prime}) - \omega \log \frac{{\bm{\pi}^*_{\bm{\rho}}}(\bm{a}^{\prime}|s^{\prime})}{{\bm{\rho}}(\bm{a}^{\prime}|s^{\prime})}  \right] \]
\fi
	
Proposition \ref{Q_coro} gives an iterative formula for $Q^*_{\bm{\rho}}(s,\bm{a})$, with which we can have a loss function and update rule for learning the critic,
\iffalse
\begin{equation}
	\mathcal{L}_{Q}=\mathbb{E}\left[\left(Q_{\phi}(s, \boldsymbol{a})-y\right)^{2}\right], %\label{Q-loss1}\\
	\text{ where } y=r(s, \boldsymbol{a})+ \gamma \left(Q_{\tilde{\phi}}\left(s^{\prime}, \boldsymbol{a}^{\prime}\right)-\omega \log \frac{\boldsymbol{\pi}\left( \boldsymbol{a}^{\prime}|s^{\prime}\right)}{\boldsymbol{\rho}\left( \boldsymbol{a}^{\prime}|s^{\prime}\right)} \right), \label{Q-loss2} \\
	%\vspace{-0.4cm}
	% 	& \phi = \phi - \alpha \nabla_\phi \mathcal{L}_Q \\
	% 	& \tilde{\phi} =(1-\tau) \tilde{\phi}+\tau \phi,  
	%\label{target-critic}
\end{equation}
\fi
\begin{gather*}
	\mathcal{L}_{Q}=\mathbb{E}\left[\left(Q_{\phi}(s, \boldsymbol{a})-y\right)^{2}\right], \\ %\label{Q-loss1}\\
	\text{ where } y=r(s, \boldsymbol{a})+ \gamma \left(Q_{\tilde{\phi}}\left(s^{\prime}, \boldsymbol{a}^{\prime}\right)-\omega \log \frac{\boldsymbol{\pi}\left( \boldsymbol{a}^{\prime}|s^{\prime}\right)}{\boldsymbol{\rho}\left( \boldsymbol{a}^{\prime}|s^{\prime}\right)} \right), \label{Q-loss2} \notag
	%\vspace{-0.4cm}
	% 	& \phi = \phi - \alpha \nabla_\phi \mathcal{L}_Q \\
	% 	& \tilde{\phi} =(1-\tau) \tilde{\phi}+\tau \phi,  
	%\label{target-critic}
\end{gather*}
where $\phi$ and $\tilde{\phi}$ are respectively the weights of Q-function and target Q-function.%, $\alpha$ is the learning rate, and $\tau$ is the hyperparameter for soft update. 
The update of Q-function is similar to that in general MDP, except that the action for next state could be chosen \textit{arbitrarily} while it must be the action that maximizes Q-function for next state in general MDP. This property greatly enhances the flexibility of learning Q-function, \textit{e.g.}, we can easily extend it to TD($\lambda$).

\begin{figure*}[t]
	\centering
	\includegraphics[width=1\textwidth]{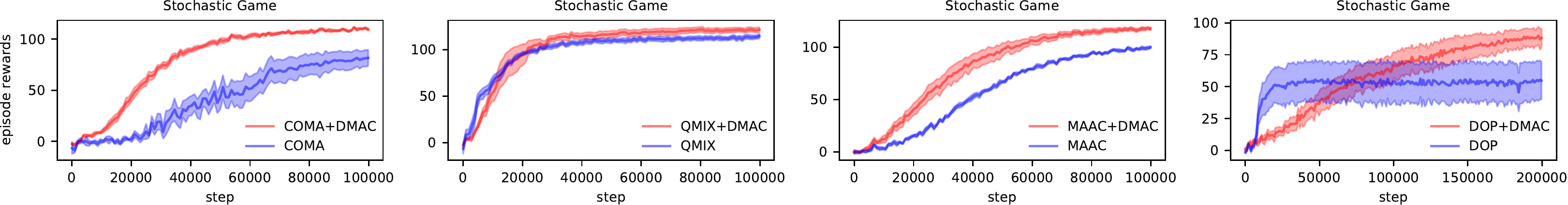}
	\vspace*{-0.7cm}
	\caption{Learning curves in terms of episode rewards of COMA, MAAC, QMIX and DOP groups in randomly generated stochastic game.}
	\label{matrix-four}
	\vspace*{-0.2cm}
\end{figure*}

\subsection{Divergence-Regularized Actors}

DAPO \citep{DIV-AUG} analyzes the divergence-regularized MDP from the perspective of policy gradient theorem \citep{PG} and gives an on-policy update rule for single-agent RL. Unlike existing work, we focus on a different perspective and derive an off-policy update rule by taking into consideration the characteristics of MARL. 

From Lemma \ref{policy-theorem}, we can obtain an optimization target for policy improvement,
\iffalse
\begin{equation}\nonumber
	%\begin{split}
	\arg  \min_{\bm{\pi}} D_{\operatorname{KL}}\left( {\bm{\pi}}(\cdot|s) \| {\bm{\rho}}(\cdot|s) \frac{\exp\left( Q^{{\bm{\pi}}}_{\bm{\rho}}(s,\cdot) / \omega \right)}{Z^{\bm{\pi}}(s)} \right) 
	= \arg \max_{\bm{\pi}} \sum_{\bm{a}} {\bm{\pi}}(\bm{a}|s) \left(  Q^{{\bm{\pi}}}_{\bm{\rho}}(s,\bm{a}) - \omega \log \frac{{\bm{\pi}}(\bm{a}|s)}{{\bm{\rho}}(\bm{a}|s)}  \right). \\
	%\end{split}
	\vspace{-0.2cm}
\end{equation}
\fi
\begin{align*}
	& \arg  \min_{\bm{\pi}} D_{\operatorname{KL}}\bigg( {\bm{\pi}}(\cdot|s) \| {\bm{\rho}}(\cdot|s) \frac{\exp\left( Q^{{\bm{\pi}}_{\operatorname{old}}}_{\bm{\rho}}(s,\cdot) / \omega \right)}{Z^{\bm{\pi}_{\operatorname{old}}}(s)} \bigg) \\
	& \quad = \arg \max_{\bm{\pi}} \sum_{\bm{a}} {\bm{\pi}}(\bm{a}|s) \left(  Q^{{\bm{\pi}}_{\operatorname{old}}}_{\bm{\rho}}(s,\bm{a}) - \omega \log \frac{{\bm{\pi}}(\bm{a}|s)}{{\bm{\rho}}(\bm{a}|s)}  \right). 
\end{align*}
Then, we can define the objective of the actors, 
\begin{equation*}
	\mathcal{L}_{\bm{\pi}} = \mathbb{E}_{s \sim \mathcal{D}}\left[ \sum_{\bm{a}} {\bm{\pi}}(\bm{a}|s) \left(  Q^{{\bm{\pi}}_{\operatorname{old}}}_{\bm{\rho}}(s,\bm{a}) - \omega \log \frac{{\bm{\pi}}(\bm{a}|s)}{{\bm{\rho}}(\bm{a}|s)}  \right) \right],
\end{equation*}
where $\mathcal{D}$ is the replay buffer. Suppose each individual policy $\pi_i$ has a corresponding parameterization $\theta_i$. 
%Suppose each individual policy $\pi_i$ has a corresponding parameterization $\theta_i$, \textit{i.e.}, $\pi_i(a_i |s, \theta_i)$. Let $\bm{\theta} = \{\theta_1,\cdots,\theta_n\}$, then the joint policy can be denoted as ${\bm{\pi}}(\bm{a} |s, \bm{\theta}) = \prod_{i=1}^{n} \pi_i(a_i | s,\theta_i )$. 
We can obtain the following policy gradient for each agent with some derivations given in Appendix \ref{sec:app-grad},
\iffalse
\begin{equation}%\nonumber
	\nabla_{\theta_i} \mathcal{L}_{\bm{\pi}} = \mathbb{E}\bigg[  \nabla_{\theta_i} \log\pi_i(a_i|s)  \bigg( Q^{{\bm{\pi}}}_{\bm{\rho}}(s,\bm{a}) 
	- \omega \log \frac{{\bm{\pi}}(\bm{a}|s)}{{\bm{\rho}}(\bm{a}|s)} - \omega  \bigg) \bigg].
\end{equation}
\fi
\iffalse
\begin{gather*}%\nonumber
	 \nabla_{\theta_i} \mathcal{L}_{\bm{\pi}} = \mathbb{E}\bigg[  \nabla_{\theta_i} \log\pi_i(a_i|s)  C^{{\bm{\pi}}}_{\bm{\rho}}(s,\bm{a}) \bigg], \\
	\text{where  } C^{{\bm{\pi}}}_{\bm{\rho}}(s,\bm{a}) = \ Q^{{\bm{\pi}}_{\operatorname{old}}}_{\bm{\rho}}(s,\bm{a}) 
	 - \omega \log \frac{{\bm{\pi}}(\bm{a}|s)}{{\bm{\rho}}(\bm{a}|s)} - \omega.  
\end{gather*}
\fi
% \begin{multline*}
% 		\nabla_{\theta_i} \mathcal{L}_{\bm{\pi}}  = \mathbb{E}\bigg[  \nabla_{\theta_i} \log\pi_i(a_i|s) \Big( Q^{{\bm{\pi}}_{\operatorname{old}}}_{\bm{\rho}}(s,\bm{a}) \\ 
% 		\qquad\qquad- \omega \log \frac{{\bm{\pi}}(\bm{a}|s)}{{\bm{\rho}}(\bm{a}|s)} - \omega \Big) \bigg].
% \end{multline*}
%\iffalse
\begin{equation*}
	\begin{split}
	    	& \nabla_{\theta_i} \mathcal{L}_{\bm{\pi}} \\
	    	& = \mathbb{E}\bigg[  \nabla_{\theta_i} \log\pi_i(a_i|s) \Big( Q^{{\bm{\pi}}_{\operatorname{old}}}_{\bm{\rho}}(s,\bm{a}) - \omega \log \frac{{\bm{\pi}}(\bm{a}|s)}{{\bm{\rho}}(\bm{a}|s)} - \omega \Big) \bigg].
	\end{split}
\end{equation*}
%\fi

We need to point out that the key to \textit{off-policy} update is that Lemma \ref{policy-theorem} does not limit the state distribution. It only requires the condition is satisfied for each state. Therefore, we can maintain a replay buffer to cover different states as much as possible, which is a common practice in off-policy learning. DAPO uses a similar formula to ours, but it obtains the formula from policy gradient theorem, which requires the state distribution of the current policy.

Further, we can add a counterfactual baseline to the gradient. First, we have the following equation about the counterfactual baseline \citep{COMA},
\begin{equation*}
	\mathbb{E}_{s \sim \mathcal{D}, \bm{a} \sim \bm{\pi}}\left[  \nabla_{\theta_i}\log{\pi}_i(a_i|s) b(s,a_{-i}) \right] = 0,
\end{equation*}
where $a_{-i}$ denotes the joint action of all agents except agent $i$. Next, we take the baseline as 
\begin{equation*}
	b(s,a_{-i}) = \mathbb{E}_{a_i \sim \pi_i} [(  Q^{{\bm{\pi}}_{\operatorname{old}}}_{\bm{\rho}}(s,\bm{a}) - \omega \log \frac{{\bm{\pi}}(\bm{a}|s)}{{\bm{\rho}}(\bm{a}|s)} - \omega)].
\end{equation*}
Then, the gradient for each agent $i$ can be modified as follows,
\iffalse
\begin{equation*}
	\begin{split}
		& \nabla_{\theta_i} \mathcal{L}_{\bm{\pi}} = \mathbb{E} [\nabla_{\theta_i} \log\pi_i(a_i|s) ( Q^{\bm{\pi}}_{\bm{\rho}}(s,\bm{a}) - \omega\log \frac{{\bm{\pi}}(\bm{a}|s)}{{\bm{\rho}}(\bm{a}|s)}  - \omega -b(s,a_{-i}) ) ] \\
		& = \mathbb{E}[  \nabla_{\theta_i} \log\pi_i(a_i|s) ( Q^{{\bm{\pi}}}_{\bm{\rho}}(s,\bm{a})  - \omega \log \frac{\pi_i(a_i|s)}{\rho_i(a_i|s)} 
		- \mathbb{E}_{a_i \sim \pi_i} [ Q^{{\bm{\pi}}}_{\bm{\rho}}(s,\bm{a})] + \omega D_{\operatorname{KL}}(\pi_i(\cdot|s) \| \rho_i(\cdot|s)) ) ]. \label{pg-final}
	\end{split}
\end{equation*}
\fi
\iffalse
\begin{equation*}
	\begin{split}
		& \nabla_{\theta_i} \mathcal{L}_{\bm{\pi}} \\
		& = \mathbb{E} \bigg[\nabla_{\theta_i} \log\pi_i(a_i|s) ( Q^{\bm{\pi}}_{\bm{\rho}}(s,\bm{a}) - \omega\log \frac{{\bm{\pi}}(\bm{a}|s)}{{\bm{\rho}}(\bm{a}|s)} \\
		& \quad - \omega -b(s,a_{-i}) ) \bigg] \\
		& = \mathbb{E} \bigg[  \nabla_{\theta_i} \log\pi_i(a_i|s) ( Q^{{\bm{\pi}}}_{\bm{\rho}}(s,\bm{a})  - \omega \log \frac{\pi_i(a_i|s)}{\rho_i(a_i|s)} \\
		& \quad - \mathbb{E}_{a_i \sim \pi_i} [ Q^{{\bm{\pi}}}_{\bm{\rho}}(s,\bm{a})] + \omega D_{\operatorname{KL}}(\pi_i(\cdot|s) \| \rho_i(\cdot|s)) ) \bigg]. \label{pg-final}
	\end{split}
\end{equation*}
\fi
%\vspace{-0.4cm}
\begin{equation*}
    %\vspace{-0.25cm}
	\begin{split}
		 \nabla_{\theta_i} \mathcal{L}_{\bm{\pi}} 
		& = \mathbb{E} \bigg[  \nabla_{\theta_i} \log\pi_i(a_i|s) ( Q^{{\bm{\pi}}_{\operatorname{old}}}_{\bm{\rho}}(s,\bm{a})  - \omega \log \frac{\pi_i(a_i|s)}{\rho_i(a_i|s)} \\
		& - \mathbb{E}_{a_i \sim \pi_i} [ Q^{{\bm{\pi}}_{\operatorname{old}}}_{\bm{\rho}}(s,\bm{a})] + \omega D_{\operatorname{KL}}(\pi_i(\cdot|s) \| \rho_i(\cdot|s)) ) \bigg]. \label{pg-final}
	\end{split}
\end{equation*}
In addition to variance reduction and credit assignment, this counterfactual baseline eliminates the policies of other agents from the gradient. This property makes it convenient to calculate the gradient and easy to select the target policy for each agent. Moreover, if the linear value decomposition condition is satisfied, we have the following gradient formula,
\iffalse
\begin{equation}
	%\begin{split}
	\nabla_{\theta_i} \mathcal{L}_{\bm{\pi}}  = \mathbb{E}[  \nabla_{\theta_i} \log\pi_i(a_i|s) ( k_i(s) A^{{\pi_i}}_{\bm{\rho}}(s,a_i)  - \omega \log \frac{\pi_i(a_i|s)}{\rho_i(a_i|s)} 
	+ \omega D_{\operatorname{KL}}(\pi_i(\cdot|s) \| \rho_i(\cdot|s)) ) ],
	%\end{split}
\end{equation}
\fi
%\vspace{-0.4cm}
\begin{multline*}
	\nabla_{\theta_i} \mathcal{L}_{\bm{\pi}}  = \mathbb{E}\bigg[  \nabla_{\theta_i} \log\pi_i(a_i|s) \big( k_i(s) A^{{\pi^i_{\operatorname{old}}}}_{\bm{\rho}}(s,a_i) \\
	\qquad - \omega \log \frac{\pi_i(a_i|s)}{\rho_i(a_i|s)} 
	+ \omega D_{\operatorname{KL}}(\pi_i(\cdot|s) \| \rho_i(\cdot|s)) \big) \bigg],
\end{multline*}
\[
\text{where } A^{\pi^i_{\operatorname{old}}}_{\bm{\rho}}(s,a_i) = Q^{\pi^i_{\operatorname{old}}}_{\bm{\rho}}(s,a_i) - \mathbb{E}_{\tilde{a}_i \sim \pi_i} \left[ Q^{\pi^i_{\operatorname{old}}}_{\bm{\rho}}(s,\tilde{a}_i)\right].
\]

\subsection{Algorithm}
% 	\vspace{-0.2cm}
% 	\begin{align}
% 		\theta_i &= \theta_i + \beta \nabla_{\theta_i} \mathcal{L}_{\bm{\pi}} \label{my-mirror} \\
% 		\tilde{\theta}_i &= (1 - \tau ) \tilde{\theta}_i + \tau \theta_i \label{target-policy},
% 		\vspace{-0.6cm}
% 	\end{align}

%Every iteration of \eqref{mirror} needs a converged policy. However, it is intractable to perform this update rule precisely in practice. Thus, we propose an alternative approximate method. For each agent, we update the policy $\pi_i$ and the target policy $\rho_i$ as  $\theta_i = \theta_i + \beta \nabla_{\theta_i} \mathcal{L}_{\bm{\pi}}$ and $\tilde{\theta}_i = (1 - \tau ) \tilde{\theta}_i + \tau \theta_i$,where $\beta$ is the learning rate, $\tilde{\theta}_i$ is the weights of $\rho_i$, and $\tau$ is the hyperparameter for soft update. Here we use one gradient step to replace the $\max$ operator in (\ref{mirror}). From Theorem \ref{policy_iteration} and previous discussion, we know that optimizing $\mathcal{L}_{\bm{\pi}}$ can maximize $J_{\bm{\rho}}({\bm{\pi}})$, so we use $\nabla_{\theta_i}\mathcal{L}_{\bm{\pi}}$ in the gradient step for off-policy training instead of the gradient step directly optimizing $J_{\bm{\rho}}({\bm{\pi}})$ in (\ref{alter-J}). Moreover, as the convergence of (\ref{MD-process}) is guaranteed only if the target policy $\bm{\rho}$ is fixed, we softly update the target policy as the moving average of the policy to prevent the instability caused by the large change of the target policy and hence obtain stable policy improvement. 
Every iteration of \eqref{mirror} needs a converged policy. However, it is intractable to perform this update rule precisely in practice. Thus, we propose an alternative approximate method. For each agent, we update the policy $\pi_i$ and the target policy $\rho_i$ respectively as  $\theta_i = \theta_i + \beta \nabla_{\theta_i} \mathcal{L}_{\bm{\pi}}$ and $\tilde{\theta}_i = (1 - \tau ) \tilde{\theta}_i + \tau \theta_i$, where $\beta$ is the learning rate, $\tilde{\theta}_i$ is the weights of $\rho_i$, and $\tau$ is the hyperparameter for soft update. Here we use one gradient step to replace the $\max$ operator in (\ref{mirror}). From Theorem \ref{policy_iteration} and previous discussion, we know that optimizing $\mathcal{L}_{\bm{\pi}}$ can maximize $J_{\bm{\rho}}({\bm{\pi}})$, so we use $\nabla_{\theta_i}\mathcal{L}_{\bm{\pi}}$ in the gradient step for off-policy training instead of the gradient step directly optimizing $J_{\bm{\rho}}({\bm{\pi}})$ in (\ref{alter-J}). 

Moreover, even if we take the target policies $\{\bm{\rho}^k\}$ as the moving average of the policies $\{\bm{\pi}^k\}$, the properties such as monotonic improvements, convergences of value functions and policies, and the bound of the biases, will be still conserved. The details of these results are included in Appendix \ref{sec:app-MA}.

Now we have all the update rules of DMAC. The training of DMAC is a typical off-policy learning process, which is given in Appendix \ref{app:algo} for completeness. 

\begin{figure}[b!]
\vspace{-0.3cm}
	\centering
	\includegraphics[width=.37\textwidth]{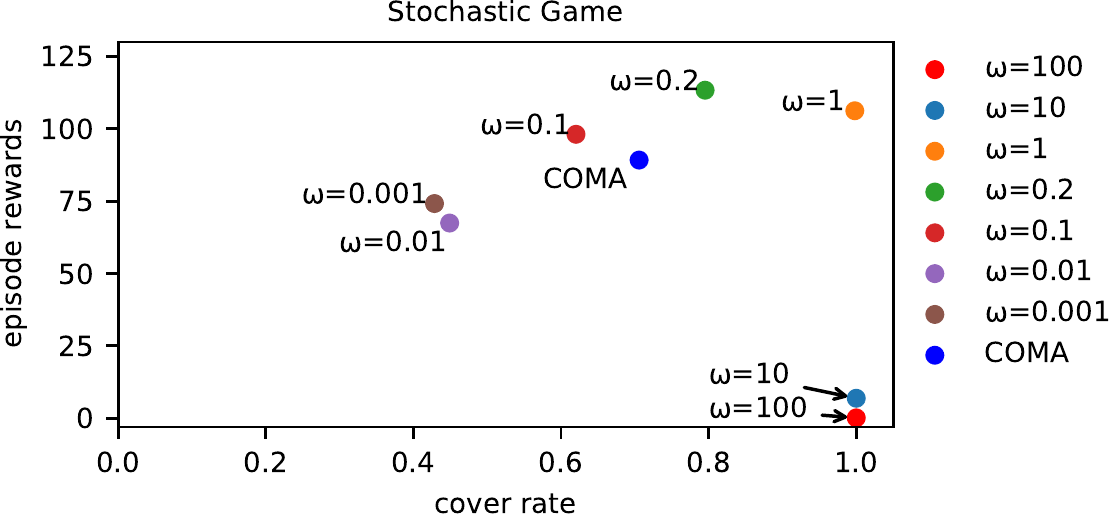}
	\vspace{-0.3cm}
	\caption{The scatter plot of the converged COMA and COMA+DMAC with different $\omega$ in terms of the episode rewards and cover rate in the randomly generated stochastic game.}
	\label{coma-cover-rate}
\end{figure}

\iffalse
\begin{wrapfigure}{htbp}{0.36\textwidth}
	\vspace{-0.4cm}
	\setlength{\abovecaptionskip}{2pt}
	\centering
	\includegraphics[width=.31\textwidth]{coma_cover_rate.pdf}
	%		\vspace{-0.6cm}
	\caption{The learning curves in terms of cover rates of COMA and COMA+DMAC in the randomly generated stochastic game. }
	\label{coma-cover-rate}
	\vspace{-0.4cm}
\end{wrapfigure}
\fi

\begin{figure*}[t]
	\centering
	%\vspace{-0.2cm}
	\setlength{\abovecaptionskip}{2pt}
	\includegraphics[width = 1\textwidth]{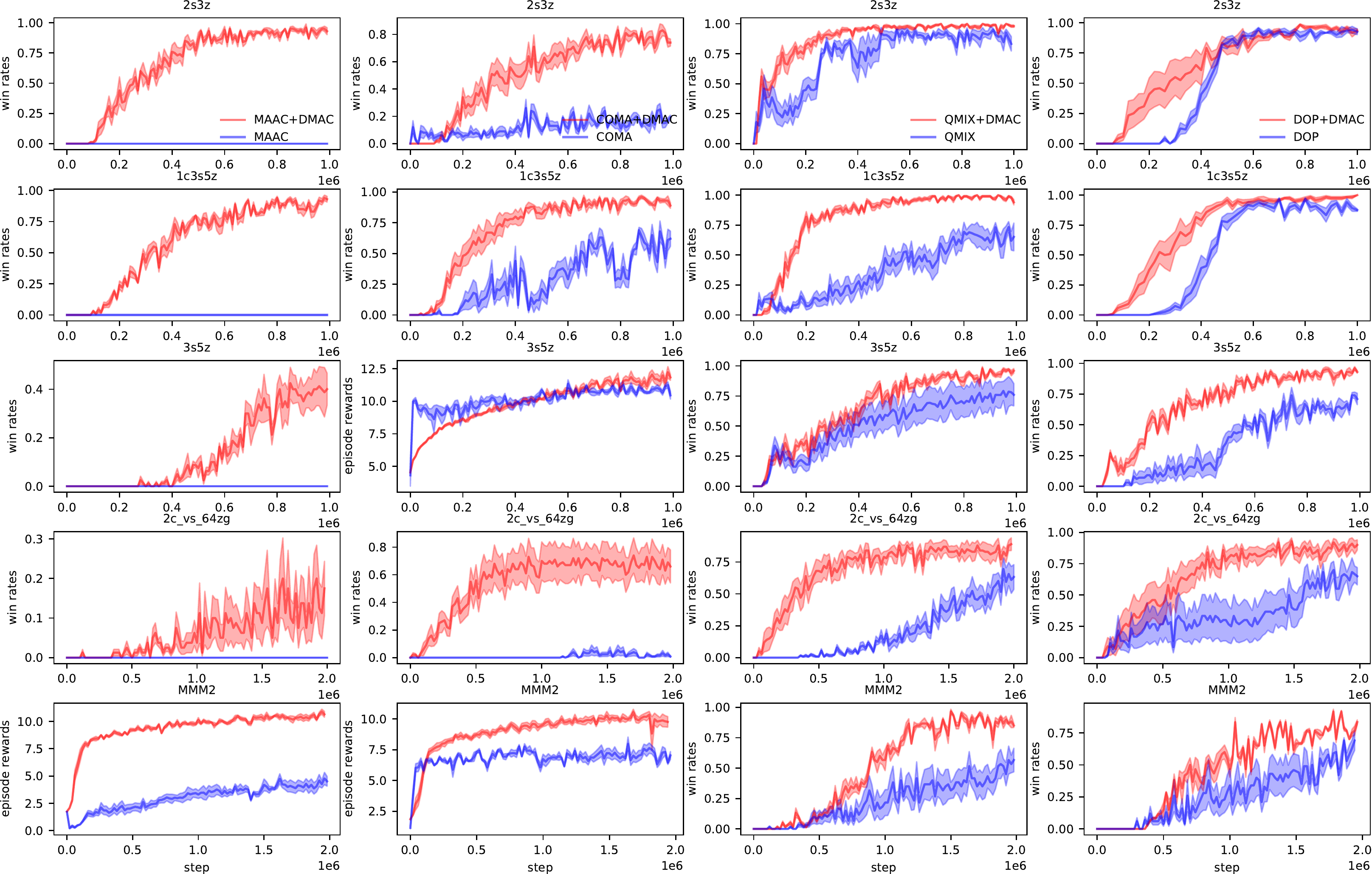}
	\vspace*{-0.7cm} 
	\caption{Learning curves in terms of win rates or episode rewards of COMA, MAAC, QMIX and DOP groups in five SMAC maps (each row corresponds to a map and each column corresponds to a group).}
	\vspace*{-0.2cm}
	\label{SMAC-curve}
\end{figure*}

\section{Experiments}

%In this section, we first show the beneficial of exploration of DMAC in a didactic stochastic game, then  empirically study the benefits of DMAC, investigate how DMAC improves the performance of existing MARL algorithms, and demonstrate the advantages of divergence regularizer over entropy regularizer in MARL. 
In this section, we first empirically study the benefits of DMAC and investigate how DMAC improves the performance of existing MARL algorithms in a didactic stochastic game and five SMAC maps. Then, we demonstrate the advantages of divergence regularizer over entropy regularizer. %in cooperative MARL.
%	DMAC is a flexible framework and can be combined with many existing MARL algorithms. In the experiments, we choose four representative algorithms for different types of methods: COMA \citep{COMA} for on-policy multi-agent policy gradients, MAAC \citep{MAAC} for off-policy multi-agent actor-critic,  QMIX \citep{QMIX} for value decomposition, DOP \citep{DOP} for the combination of value decomposition and policy gradient. These algorithms need minor modifications to fit the framework of DMAC. We denote these modified algorithms as COMA+DMAC, MAAC+DMAC,  QMIX+DMAC and DOP+DMAC. Generally, our modification is limited and tries to keep the original architecture so as to fairly demonstrate the improvement of DMAC. The details of the modifications are included in Appendix \ref{app:modifications}. More details about hyperparameters are available in Appendix~\ref{app:details}.  All the curves in our plots corresponds to the mean value of 5 training runs with different random seeds, and shaded regions indicate 95\% confidence interval.

\subsection{Improvements of Existing Methods}
DMAC is a flexible framework and can be combined with many existing MARL algorithms. In the experiments, we choose four representative algorithms for different types of methods: COMA \citep{COMA} for on-policy multi-agent policy gradients, MAAC \citep{MAAC} for off-policy multi-agent actor-critic, QMIX \citep{QMIX} for value decomposition, DOP \citep{DOP} for the combination of value decomposition and policy gradient. These algorithms need minor modifications to fit the framework of DMAC. We denote these modified algorithms as COMA+DMAC, MAAC+DMAC,  QMIX+DMAC, and DOP+DMAC. Generally, our modification is limited and tries to keep the original architecture so as to fairly demonstrate the improvement of DMAC. The details of the modifications and hyperparameters are included in Appendix \ref{app:details}. 
%More details about hyperparameters are available in Appendix~\ref{app:details}. 
All the curves in our plots correspond to the mean value of five training runs with different random seeds, and shaded regions indicate 95\% confidence interval.

\subsubsection*{A Didactic Example}

We first test the four groups of methods in a stochastic game where agents share the reward. The stochastic game is generated randomly for the reward function and transition probabilities with 30 states, 3 agents and 5 actions for each agent. Each episode contains 30 timesteps in this game. The performance of these methods is illustrated in Figure \ref{matrix-four}.  We can find that DMAC performs better than the baseline at the end of the training in all the four groups. Moreover, COMA+DMAC, QMIX+DMAC and MAAC+DMAC learn faster than their baselines. Though DOP learns faster than DOP+DMAC at the start, it falls into a sub-optimal policy and DOP+DMAC finds a better policy in the end.

\begin{figure*}[t]
	\centering
	\includegraphics[width=1\textwidth]{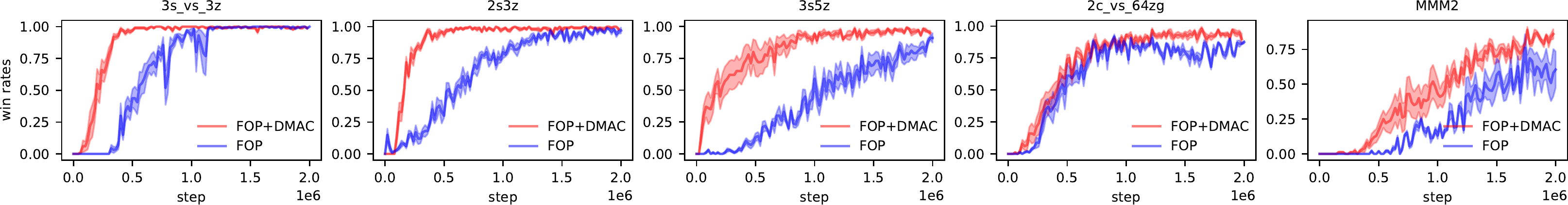}
	\vspace*{-0.7cm}
	\caption{Learning curves in terms of win rates of FOP and FOP+DMAC in five SMAC maps. }
	\label{fop-sc2}
	\vspace{-0.2cm}
\end{figure*}

%\textbf{Exploration and $\bm{\omega}$.}
We also show the benefit of exploration in this stochastic game for the convenience of statistics.  We evaluate the exploration in terms of the \textit{cover rate} of all state-action pairs, \textit{i.e.}, the ratio of the explored state-action pairs to all state-action pairs. The cover rates of COMA and COMA+DMAC with different $\omega$ are illustrated in Figure \ref{coma-cover-rate}. We first study the  effect of $\omega$ for the tradeoff between exploration and exploitation. From our previous analysis, a smaller $\omega$ gives a tighter bound for the converged performance of DMAC, but it also means that the benefits of the regularizer for exploration is smaller which will also affect the performance. So we need to choose a proper $\omega$ in practice. We can find in Figure \ref{coma-cover-rate} that when $\omega$ is large such as the case $\omega = 10$ and $\omega =  100$, the exploration is guaranteed but the performance is quite low since the regularizer is much larger than the environmental reward, and when $\omega$ is small such as the case $\omega = 0.001$ and $\omega =  0.01$, the performance is limited by the ability of exploration. Only when $\omega$ lies in a proper interval such as $\omega = 0.1$, $\omega = 0.2$ and $\omega = 1$, the agents of DMAC can obtain a good balance between exploration and exploitation in this task.

We use COMA here as a representation of the traditional policy gradient method in cooperative MARL. In practice, we choose the $\omega=0.2$ for COMA+DMAC.  We can find that the cover rate of COMA+DMAC ($\omega=0.2$) is higher than COMA, which can be an evidence for the benefit of exploration of DMAC. The cover rates of other three groups of algorithms and the learning curves for methods in Figure \ref{coma-cover-rate} are available in Appendix~\ref{app:add}.
\subsubsection*{SMAC}
%As for SMAC tasks, we evaluate 20 episodes for every 5000 training steps in the one million steps training procedure.  We select actions with the same mode as the MPE tasks in evaluation.
We test all the methods in five tasks of SMAC \citep{SMAC}. The introduction of the SMAC environment and the training details are included in Appendix \ref{app:details}. The learning curves in terms of win rate (or episode rewards) of all the methods in the five SMAC maps are illustrated in Figure \ref{SMAC-curve} (four columns for four groups of algorithms and five rows for five different maps in SMAC).  For the case that both the baseline and DMAC can hardly win such as 3s5z and MMM2 for the COMA group and MMM2 for the MAAC group, we use the episode rewards to show the difference. In addition, the learning curves in terms of episode rewards are available in Appendix~\ref{app:add}, including two additional maps, 3m and 8m. We show the empirical result of DAPO in the map of 3m in the first figure of the second column in Figure \ref{more-SMAC-curve} in Appendix~\ref{app:add}. It can be seen that DAPO cannot obtain a good performance in the simple task of SMAC, so we skip it in other SMAC tasks. The reason for the low performance of DAPO may be that DAPO omits the correction for $d_\pi(s)/d_{\pi_t}(s)$ in policy update which introduces bias in the gradient of policy, and uses V-trace as off-policy correction which however is biased. These drawbacks may be magnified in MARL settings. The superiority of our naturally off-policy method over the biased off-policy correction method can be partly seen from the large performance gap between COMA+DMAC and DAPO.

In all the five tasks, MAAC+DMAC outperforms MAAC significantly, but MAAC+DMAC does not change the network architecture of MAAC, which shows the benefits of divergence regularizer.
As for the result of COMA and COMA+DMAC. We find that COMA+DMAC has higher win rates than COMA in most cases at the end of the training, which can be attributed to the benefits of off-policy learning and exploration of divergence regularizer. Though in some cases COMA learns faster than COMA+DMAC, it falls into sub-optimal in the end. This phenomenon can be observed more clearly in the plots of episode rewards in the hard tasks like 3s5z. This can be an evidence for the advantage of divergence regularizer which helps the agents find a better policy. 

The stable policy improvement of divergence regularizer can be manifested by the small variance of the learning curves especially in the comparison between QMIX and QMIX+DMAC. In most tasks, we find that QMIX+DMAC learns substantially faster than QMIX and gets higher win rates in harder tasks.  
The results of DOP groups are illustrated in the fourth column of Figure \ref{SMAC-curve}. DOP+DMAC learns faster than DOP in most cases and finally obtains a better performance. The difference of DOP and DOP+DMAC can also partly show the advantage of naturally off-policy method to the off-policy correction method, as DOP+DMAC replaces the tree backup loss with off-policy TD($\lambda$).

DMAC improves the performance and/or convergence speed of the evaluated algorithms in most tasks. This empirically demonstrates the benefits of divergence regularizer. Moreover, the superiority of our naturally off-policy learning over the biased off-policy correction method can be partly witnessed from the empirical results.

\subsection{Comparison with Entropy Regularization}

% 	\begin{wrapfigure}{!b}{0.75\textwidth}
% 		\vspace{-0.6cm}
% 		\setlength{\abovecaptionskip}{2pt}
% 		\centering
% 		\includegraphics[width=.75\textwidth]{fop_sc2_win_rates.pdf}
% 		\vspace{-0.4cm}
% 		\caption{The learning curves of win rates of FOP+DMAC and FOP in three SMAC maps. }
% 		\label{fop-sc2}
% 		\vspace{-0.4cm}
% 	\end{wrapfigure}

FOP \citep{FOP} combines value decomposition with entropy regularization, which obtained the state-of-the-art performance in SMAC tasks. FOP has a well-tuned scheme for the temperature parameter of the entropy, so we take FOP as a strong baseline for entropy-regularized methods in cooperative MARL. We compare FOP and FOP+DMAC in five SMAC tasks, 3s\_vs\_3z, 2s3z, 3s5z, 2c\_vs\_64zg and MMM2, which respectively correspond to the three levels of difficulties (\textit{i.e.}, easy, hard, and super hard) for SMAC tasks. Three of these tasks are taken from the original paper of FOP. The modifications of FOP+DMAC are also included in Appendix \ref{app:details}. The win rates of FOP and FOP+DMAC are illustrated in Figure \ref{fop-sc2}. We can find that FOP+DMAC learns much faster than FOP in 3s\_vs\_3z, while it performs better than FOP in other four tasks. 
%the win rates of FOP and FOP+DMAC are illustrated in Figure \ref{fop-sc2}. We could find that FOP+DMAC performs better than FOP in all the three tasks. 
These results can be an evidence for the advantages of DMAC which could guarantee the monotonic improvement in the original MDP. 

\section{Conclusion}
%Unlike entropy regularizer, the divergence regularizer will not bias the converged policy.
We propose a multi-agent actor-critic framework, DMAC, for cooperative MARL. We investigate divergence regularization, derive divergence policy iteration, and deduce the update rules for the critic, policy, and target policy in multi-agent settings. DMAC is a naturally off-policy framework and the divergence regularizer is beneficial to exploration and stable policy improvement.  DMAC is also a flexible framework and can be combined with many existing MARL algorithms. It is empirically demonstrated that combining DMAC with existing MARL algorithms can improve the performance and convergence speed in a stochastic game and SMAC tasks.

\section*{Acknowledgements}
We would like to thank the anonymous reviewers for their useful comments to improve our work. This work is supported in part by NSF China under grant 61872009 and Huawei.

\bibliography{mypaper2}

\begin{thebibliography}{34}
\providecommand{\natexlab}[1]{#1}
\providecommand{\url}[1]{\texttt{#1}}
\expandafter\ifx\csname urlstyle\endcsname\relax
  \providecommand{\doi}[1]{doi: #1}\else
  \providecommand{\doi}{doi: \begingroup \urlstyle{rm}\Url}\fi

\bibitem[Agarwal et~al.(2020)Agarwal, Kumar, Sycara, and Lewis]{transfer}
Agarwal, A., Kumar, S., Sycara, K., and Lewis, M.
\newblock {Learning} {Transferable} {Cooperative} {Behavior} in {Multi}-{Agent}
  {Team}.
\newblock In \emph{International Conference on Autonomous Agents and Multiagent
  Systems (AAMAS)}, 2020.

\bibitem[Espeholt et~al.(2018)Espeholt, Soyer, Munos, Simonyan, Mnih, Ward,
  Doron, Firoiu, Harley, Dunning, et~al.]{V-trace}
Espeholt, L., Soyer, H., Munos, R., Simonyan, K., Mnih, V., Ward, T., Doron,
  Y., Firoiu, V., Harley, T., Dunning, I., et~al.
\newblock {Impala}: {Scalable} {Distributed} {Deep}-{Rl} with {Importance}
  {Weighted} {Actor}-{Learner} {Architectures}.
\newblock In \emph{International Conference on Machine Learning (ICML)}, 2018.

\bibitem[Eysenbach \& Levine(2019)Eysenbach and Levine]{whatisquestion}
Eysenbach, B. and Levine, S.
\newblock {If} {MaxEnt} {RL} {Is} {The} {Answer}, {What} {Is} {The} {Question}?
\newblock \emph{arXiv preprint arXiv:1910.01913}, 2019.

\bibitem[Foerster et~al.(2018)Foerster, Farquhar, Afouras, Nardelli, and
  Whiteson]{COMA}
Foerster, J.~N., Farquhar, G., Afouras, T., Nardelli, N., and Whiteson, S.
\newblock {Counterfactual} {Multi}-{Agent} {Policy} {Gradients}.
\newblock In \emph{AAAI Conference on Artificial Intelligence (AAAI)}, 2018.

\bibitem[Grau-Moya et~al.(2019)Grau-Moya, Leibfried, and Vrancx]{MIRL}
Grau-Moya, J., Leibfried, F., and Vrancx, P.
\newblock {Soft} {Q}-{Learning} with {Mutual}-{Information} {Regularization}.
\newblock In \emph{International Conference on Learning Representations
  (ICLR)}, 2019.

\bibitem[Haarnoja et~al.(2017)Haarnoja, Tang, Abbeel, and Levine]{deep_energy}
Haarnoja, T., Tang, H., Abbeel, P., and Levine, S.
\newblock {Reinforcement} {Learning} with {Deep} {Energy}-{Based} {Policies}.
\newblock In \emph{International Conference on Machine Learning (ICML)}, 2017.

\bibitem[Haarnoja et~al.(2018)Haarnoja, Zhou, Abbeel, and Levine]{SAC}
Haarnoja, T., Zhou, A., Abbeel, P., and Levine, S.
\newblock {Soft} {Actor}-{Critic}: {Off}-{Policy} {Maximum} {Entropy} {Deep}
  {Reinforcement} {Learning} with {A} {Stochastic} {Actor}.
\newblock In \emph{International Conference on Machine Learning (ICML)}, 2018.

\bibitem[Iqbal \& Sha(2019)Iqbal and Sha]{MAAC}
Iqbal, S. and Sha, F.
\newblock {Actor}-{Attention}-{Critic} for {Multi}-{Agent} {Reinforcement}
  {Learning}.
\newblock In \emph{International Conference on Machine Learning (ICML)}, 2019.

\bibitem[Jiang et~al.(2020)Jiang, Dun, Huang, and Lu]{DGN}
Jiang, J., Dun, C., Huang, T., and Lu, Z.
\newblock {Graph} {Convolutional} {Reinforcement} {Learning}.
\newblock In \emph{International Conference on Learning Representations
  (ICLR)}, 2020.

\bibitem[Kakade \& Langford(2002)Kakade and Langford]{CPI}
Kakade, S. and Langford, J.
\newblock {Approximately} {Optimal} {Approximate} {Reinforcement} {Learning}.
\newblock In \emph{International Conference on Machine Learning (ICML)}, 2002.

\bibitem[Lowe et~al.(2017)Lowe, Wu, Tamar, Harb, Abbeel, and Mordatch]{MADDPG}
Lowe, R., Wu, Y.~I., Tamar, A., Harb, J., Abbeel, O.~P., and Mordatch, I.
\newblock {Multi}-{Agent} {Actor}-{Critic} for {Mixed}
  {Cooperative}-{Competitive} {Environments}.
\newblock In \emph{Advances in Neural Information Processing Systems
  (NeurIPS)}, 2017.

\bibitem[Nachum et~al.(2017)Nachum, Norouzi, Xu, and Schuurmans]{SAC-theory}
Nachum, O., Norouzi, M., Xu, K., and Schuurmans, D.
\newblock {Bridging} {The} {Gap} {Between} {Value} and {Policy} {Based}
  {Reinforcement} {Learning}.
\newblock In \emph{Advances in Neural Information Processing Systems
  (NeurIPS)}, 2017.

\bibitem[Nachum et~al.(2018)Nachum, Norouzi, Xu, and Schuurmans]{Trust-PCL}
Nachum, O., Norouzi, M., Xu, K., and Schuurmans, D.
\newblock {Trust}-{Pcl}: {An} {Off}-{Policy} {Trust} {Region} {Method} for
  {Continuous} {Control}.
\newblock In \emph{International Conference on Learning Representations
  (ICLR)}, 2018.

\bibitem[Neu et~al.(2017)Neu, Jonsson, and G{\'o}mez]{unifiedview}
Neu, G., Jonsson, A., and G{\'o}mez, V.
\newblock {A} {Unified} {View} of {Entropy}-{Regularized} {Markov} {Decision}
  {Processes}.
\newblock \emph{arXiv preprint arXiv:1705.07798}, 2017.

\bibitem[Oliehoek et~al.(2016)Oliehoek, Amato, et~al.]{Dec-POMDP}
Oliehoek, F.~A., Amato, C., et~al.
\newblock \emph{{A} {Concise} {Introduction} {To} {Decentralized} {POMDPs}},
  volume~1.
\newblock Springer, 2016.

\bibitem[Pan et~al.(2021)Pan, Rashid, Peng, Huang, and Whiteson]{REQMIX}
Pan, L., Rashid, T., Peng, B., Huang, L., and Whiteson, S.
\newblock {Regularized} {Softmax} {Deep} {Multi}-{Agent} {Q}-{Learning}.
\newblock \emph{Advances in Neural Information Processing Systems (NeurIPS)},
  2021.

\bibitem[Peng et~al.(2021)Peng, Rashid, de~Witt, Kamienny, Torr, B{\"o}hmer,
  and Whiteson]{FACMAC}
Peng, B., Rashid, T., de~Witt, C. A.~S., Kamienny, P.-A., Torr, P.~H.,
  B{\"o}hmer, W., and Whiteson, S.
\newblock {FACMAC}: {Factored} {Multi}-{Agent} {Centralised} {Policy}
  {Gradients}.
\newblock In \emph{Advances in Neural Information Processing Systems
  (NeurIPS)}, 2021.

\bibitem[Rashid et~al.(2018)Rashid, Samvelyan, De~Witt, Farquhar, Foerster, and
  Whiteson]{QMIX}
Rashid, T., Samvelyan, M., De~Witt, C.~S., Farquhar, G., Foerster, J., and
  Whiteson, S.
\newblock {QMIX}: {Monotonic} {Value} {Function} {Factorisation} for {Deep}
  {Multi}-{Agent} {Reinforcement} {Learning}.
\newblock In \emph{International Conference on Machine Learning (ICML)}, 2018.

\bibitem[Samvelyan et~al.(2019)Samvelyan, Rashid, de~Witt, Farquhar, Nardelli,
  Rudner, Hung, Torr, Foerster, and Whiteson]{SMAC}
Samvelyan, M., Rashid, T., de~Witt, C.~S., Farquhar, G., Nardelli, N., Rudner,
  T.~G., Hung, C.-M., Torr, P.~H., Foerster, J.~N., and Whiteson, S.
\newblock {The} {StarCraft} {Multi}-{Agent} {Challenge}.
\newblock In \emph{AAMAS}, 2019.

\bibitem[Schulman et~al.(2015)Schulman, Levine, Abbeel, Jordan, and
  Moritz]{TRPO}
Schulman, J., Levine, S., Abbeel, P., Jordan, M., and Moritz, P.
\newblock {Trust} {Region} {Policy} {Optimization}.
\newblock In \emph{International Conference on Machine Learning (ICML)}, 2015.

\bibitem[Schulman et~al.(2017)Schulman, Wolski, Dhariwal, Radford, and
  Klimov]{PPO}
Schulman, J., Wolski, F., Dhariwal, P., Radford, A., and Klimov, O.
\newblock {Proximal} {Policy} {Optimization} {Algorithms}.
\newblock \emph{arXiv preprint arXiv:1707.06347}, 2017.

\bibitem[{Son} et~al.(2019){Son}, {Kim}, {Kang}, {Hostallero}, and {Yi}]{QTRAN}
{Son}, K., {Kim}, D., {Kang}, W.~J., {Hostallero}, D.~E., and {Yi}, Y.
\newblock {QTRAN}: {Learning} {To} {Factorize} with {Transformation} for
  {Cooperative} {Multi}-{Agent} {Reinforcement} {Learning}.
\newblock In \emph{International Conference on Machine Learning (ICML)}, 2019.

\bibitem[Sun et~al.(2021)Sun, Lee, and Lee]{DFAC}
Sun, W.-F., Lee, C.-K., and Lee, C.-Y.
\newblock {DFAC} {Framework}: {Factorizing} {The} {Value} {Function} {Via}
  {Quantile} {Mixture} for {Multi}-{Agent} {Distributional} {Q}-{Learning}.
\newblock In \emph{International Conference on Machine Learning (ICML)}, 2021.

\bibitem[Sunehag et~al.(2018)Sunehag, Lever, Gruslys, Czarnecki, Zambaldi,
  Jaderberg, Lanctot, Sonnerat, Leibo, Tuyls, et~al.]{VDN}
Sunehag, P., Lever, G., Gruslys, A., Czarnecki, W.~M., Zambaldi, V.~F.,
  Jaderberg, M., Lanctot, M., Sonnerat, N., Leibo, J.~Z., Tuyls, K., et~al.
\newblock {Value}-{Decomposition} {Networks} {For} {Cooperative}
  {Multi}-{Agent} {Learning} {Based} {On} {Team} {Reward}.
\newblock In \emph{International Conference on Autonomous Agents and MultiAgent
  Systems (AAMAS)}, 2018.

\bibitem[Sutton \& Barto(2018)Sutton and Barto]{RLBOOK}
Sutton, R.~S. and Barto, A.~G.
\newblock \emph{{Reinforcement} {Learning}: {An} {Introduction}}.
\newblock MIT press, 2018.

\bibitem[Sutton et~al.(2000)Sutton, McAllester, Singh, and Mansour]{PG}
Sutton, R.~S., McAllester, D.~A., Singh, S.~P., and Mansour, Y.
\newblock {Policy} {Gradient} {Methods} for {Reinforcement} {Learning} with
  {Function} {Approximation}.
\newblock In \emph{Advances in Neural Information Processing Systems
  (NeurIPS)}, 2000.

\bibitem[Touati et~al.(2020)Touati, Zhang, Pineau, and Vincent]{stable-policy}
Touati, A., Zhang, A., Pineau, J., and Vincent, P.
\newblock {Stable} {Policy} {Optimization} {Via} {Off}-{Policy} {Divergence}
  {Regularization}.
\newblock In \emph{Conference on Uncertainty in Artificial Intelligence (UAI)},
  2020.

\bibitem[Wang et~al.(2021{\natexlab{a}})Wang, Ren, Liu, Yu, and Zhang]{QPLEX}
Wang, J., Ren, Z., Liu, T., Yu, Y., and Zhang, C.
\newblock {QPLEX}: {Duplex} {Dueling} {Multi}-{Agent} {Q}-{Learning}.
\newblock In \emph{International Conference on Learning Representations
  (ICLR)}, 2021{\natexlab{a}}.

\bibitem[Wang et~al.(2019)Wang, Li, Xiong, and Zhang]{DIV-AUG}
Wang, Q., Li, Y., Xiong, J., and Zhang, T.
\newblock {Divergence}-{Augmented} {Policy} {Optimization}.
\newblock In \emph{Advances in Neural Information Processing Systems
  (NeurIPS)}, 2019.

\bibitem[Wang et~al.(2021{\natexlab{b}})Wang, Han, Wang, Dong, and Zhang]{DOP}
Wang, Y., Han, B., Wang, T., Dong, H., and Zhang, C.
\newblock {DOP}: {Off}-{Policy} {Multi}-{Agent} {Decomposed} {Policy}
  {Gradients}.
\newblock In \emph{International Conference on Learning Representations
  (ICLR)}, 2021{\natexlab{b}}.

\bibitem[Yang et~al.(2019)Yang, Li, and Zhang]{regularizer}
Yang, W., Li, X., and Zhang, Z.
\newblock {A} {Regularized} {Approach} {To} {Sparse} {Optimal} {Policy} in
  {Reinforcement} {Learning}.
\newblock In \emph{Advances in Neural Information Processing Systems
  (NeurIPS)}, 2019.

\bibitem[Yang et~al.(2020)Yang, Hao, Liao, Shao, Chen, Liu, and Tang]{Qatten}
Yang, Y., Hao, J., Liao, B., Shao, K., Chen, G., Liu, W., and Tang, H.
\newblock {Qatten}: {A} {General} {Framework} for {Cooperative} {Multiagent}
  {Reinforcement} {Learning}.
\newblock \emph{arXiv preprint arXiv:2002.03939}, 2020.

\bibitem[Yu et~al.(2021)Yu, Velu, Vinitsky, Wang, Bayen, and Wu]{MAPPO}
Yu, C., Velu, A., Vinitsky, E., Wang, Y., Bayen, A.~M., and Wu, Y.
\newblock {The} {Surprising} {Effectiveness} of {PPO} in {Cooperative},
  {Multi}-{Agent} {Games}.
\newblock \emph{arXiv preprint arXiv:2103.01955}, 2021.

\bibitem[Zhang et~al.(2021)Zhang, Li, Wang, Xie, and Lu]{FOP}
Zhang, T., Li, Y., Wang, C., Xie, G., and Lu, Z.
\newblock {FOP}: {Factorizing} {Optimal} {Joint} {Policy} of
  {Maximum}-{Entropy} {Multi}-{Agent} {Reinforcement} {Learning}.
\newblock In \emph{International Conference on Machine Learning (ICML)}, 2021.

\end{thebibliography}
\bibliographystyle{icml2022}

\newpage
\appendix
\onecolumn

\section{Proofs}
	\label{app:proofs}

\subsection{Proposition \ref{converged_policy}}\label{app:converged_policy}

	\begin{proof}
	    For the sake of simplicity, we define $\tilde{Q}^{k} = Q^*_{\bm{\pi}^{k - 1}}$ and $\tilde{V}^{k} = V^*_{\bm{\pi}^{k - 1}}$. From the definition of iteration \eqref{mirror}, we have $\bm{\pi}^k = \bm{\pi}^*_{\bm{\pi}^{k-1}}$. Then from Proposition \ref{monotonic_converge}, we also have $\tilde{Q}^{k} \to \tilde{Q}^{*}$ and $\tilde{V}^{k} \to \tilde{V}^{*}$.
	    
	    From Proposition \ref{theorem1}, we have $\bm{\pi}^k(\bm{a}|s) \propto \bm{\pi}^{k-1}(\bm{a}|s) \exp(\frac{\tilde{Q}^k(s,\bm{a})}{\omega})$. We define $f^k(s,\bm{a}) = \exp(\frac{\tilde{Q}^k(s,\bm{a})}{\omega})$ and $f^*(s,\bm{a}) = \exp(\frac{\tilde{Q}^*(s,\bm{a})}{\omega})$, then we have $f^k(s,\bm{a}) \to f^*(s,\bm{a})$. Next we will show that $\bm{\pi}^k(\bm{a}|s) \propto \bm{\pi}^{0}(\bm{a}|s) \Pi_{i = 1}^k f^i(s,\bm{a}) $ or $\bm{\pi}^k(\bm{a}|s) = \frac{\bm{\pi}^{0}(\bm{a}|s) \Pi_{i = 1}^k f^i(s,\bm{a})}{Z^k(s)}$ by induction, where $Z^k(s) = \sum_{\bm{a}^\prime} \bm{\pi}^{0}(\bm{a}^\prime|s) \Pi_{i = 1}^k f^i(s,\bm{a}^\prime) $ . 
	    \begin{align*}
	        \bm{\pi}^k(\bm{a}|s) & = \frac{\bm{\pi}^{k-1}(\bm{a}|s)f^k(s,\bm{a})}{\sum_{\bm{a}^\prime}\bm{\pi}^{k-1}(\bm{a}^\prime|s)f^k(s,\bm{a}^\prime) } \\
	        & = \frac{  \frac{\bm{\pi}^{0}(\bm{a}|s) \Pi_{i = 1}^{k-1} f^i(s,\bm{a})}{Z^{k -1}(s)}f^k(s,\bm{a})}{\sum_{\bm{a}^\prime}\frac{\bm{\pi}^{0}(\bm{a}^\prime|s) \Pi_{i = 1}^{k-1} f^i(s,\bm{a}^\prime)}{Z^{k -1}(s)}f^k(s,\bm{a}^\prime) } \quad \text{(By induction)} \\
	        & = \frac{  \bm{\pi}^{0}(\bm{a}|s) \Pi_{i = 1}^{k}f^i(s,\bm{a}) }{\sum_{\bm{a}^\prime} \bm{\pi}^{0}(\bm{a}^\prime|s) \Pi_{i = 1}^{k}f^i(s,\bm{a}^\prime) } \\
	        & = \frac{ \bm{\pi}^{0}(\bm{a}|s) \Pi_{i = 1}^{k}f^i(s,\bm{a})}{Z^k(s)}
	    \end{align*}
	    
	    We define $f^*_s = \max_{\bm{a}} f^*(s,\bm{a})$, then we will consider $\tilde{\pi}^*(\bm{a}|s) = \lim_{k \to \infty} \pi^k(\bm{a}|s)$.
	    \begin{align*}
	        \lim_{k \to \infty} \pi^k(\bm{a}|s) & = \lim_{k \to \infty} 
	        \frac{  
	        \bm{\pi}^{0}(\bm{a}|s) \Pi_{i = 1}^{k}f^i(s,\bm{a}) 
	        }
	        {
	        \sum_{\bm{a}^\prime} \bm{\pi}^{0}(\bm{a}^\prime|s) \Pi_{i = 1}^{k}f^i(s,\bm{a}^\prime) 
	        } \\
	        & = \lim_{k \to \infty}
	        \frac{  
	        \bm{\pi}^{0}(\bm{a}|s) 
	            \frac{
    	            \Pi_{i = 1}^{k}f^i(s,\bm{a}) }{f^*(s,\bm{a})^k}
    	       \frac{f^*(s,\bm{a})^k}{(f^*_s)^k}
	        }
	        {
	        \sum_{\bm{a}^\prime} \bm{\pi}^{0}(\bm{a}^\prime|s) 
	            \frac{
    	            \Pi_{i = 1}^{k}f^i(s,\bm{a}^\prime) }{f^*(s,\bm{a}^\prime)^k}
    	       \frac{f^*(s,\bm{a}^\prime)^k}{(f^*_s)^k}
	        } \\
	        & = \frac
	            {
	                \bm{\pi}^{0}(\bm{a}|s)
	                 \lim_{k \to \infty}
	                \frac{
    	            \Pi_{i = 1}^{k}f^i(s,\bm{a}) }{f^*(s,\bm{a})^k}
    	            \frac{f^*(s,\bm{a})^k}{(f^*_s)^k}
	            }
	            {
	                \sum_{\bm{a}^\prime}
	                \bm{\pi}^{0}(\bm{a}^\prime|s)
	                 \lim_{k \to \infty}
	                \frac{
    	            \Pi_{i = 1}^{k}f^i(s,\bm{a}^\prime) }{f^*(s,\bm{a}^\prime)^k}
    	            \frac{f^*(s,\bm{a^\prime})^k}{(f^*_s)^k}
	            } \\
	       & = \mathbbm{1}{(\bm{a} \in U_s)} \frac{\bm{\pi}^0(\bm{a}|s)}{\sum_{\bm{a}^\prime \in U_s}\bm{\pi}^0(\bm{a}^\prime|s)}
	    \end{align*}
	    The last equation is actually from two simple conclusion: (1) If a sequence $\{a_k\} >0 $ and $a_k \to A > 0$, then $\lim_{n \to \infty} \frac{\Pi_{k = 1}^na_k}{A^n} = 1$ ; (2) For $A > B > 0$,  $\lim_{n \to \infty} \frac{B^n}{A^n} = 0$. So $\lim_{k \to \infty}
	                \frac{
    	            \Pi_{i = 1}^{k}f^i(s,\bm{a}) }{f^*(s,\bm{a})^k}
    	            \frac{f^*(s,\bm{a})^k}{(f^*_s)^k} = \mathbbm{1}{(\bm{a} \in U_s)} $. 
	\end{proof}

		\subsection{Theorem \ref{theorem:global_optimal}}
	\label{app:global_optimal}
	We define two operator $\Gamma$ and $\Gamma_\omega$ as 
	\begin{align*}
	    &\Gamma V(s) = \max_{\bm{\pi}} \sum_{\bm{a}} \pi(\bm{a}|s) (r(s,\bm{a}) + \gamma \mathbb{E}[V(s^\prime)]) \\
	    & \Gamma_\omega V(s) = \max_{\bm{\pi}} \sum_{\bm{a}} \pi(\bm{a}|s) (r(s,\bm{a}) + \gamma \mathbb{E}[V(s^\prime)] ) - \omega D_{\operatorname{KL}}(\bm{\pi}(\cdot|s) || \tilde{\bm{\pi}}^*(\cdot|s) )  
	\end{align*}
	From the result of traditional Q-learning \citep{RLBOOK}, we know that $\Gamma$ is a $\gamma$-contraction and the unique fixed point is the global optimal V-function $V^*$. 
	
	As for $\Gamma_\omega$, we have the following lemma from \citet{regularizer}:
	\begin{lemma} \label{reg-lemma}
	\mbox{}\par
	    \begin{itemize}
	        \item[$(1)$] $\Gamma_\omega$ is a $\gamma$-contraction.
	        \item[$(2)$] If $V_1(s) \le V_2(s)$ for all state $s$, then $\Gamma_\omega V_1(s) \le \Gamma_\omega V_2(s)$ for all state $s$
	        \item[$(3)$] For any constant $c$, we define $(V+c)(s) = V(s) + c $, then $\Gamma_\omega (V+c)(s) = \Gamma_\omega V(s) + \gamma c$
	    \end{itemize}
	\end{lemma}
	Moreover, from the definition we know that $\tilde{V}^k(s) = \max_{\bm{\pi}} \sum_{\bm{a}} \pi(\bm{a}|s) (r(s,\bm{a}) + \gamma \mathbb{E}[\tilde{V}^k(s^\prime)] )- \omega D_{\operatorname{KL}}(\bm{\pi}(\cdot|s) || {\bm{\pi}}^{k - 1}(\cdot|s) $ and $\tilde{V}^k \to \tilde{V}^*,\bm{\pi}^{k} \to \tilde{\bm{\pi}}^*$, then we have:
	\begin{align*}
	    \tilde{V}^*(s) = \max_{\bm{\pi}} \sum_{\bm{a}} \pi(\bm{a}|s) (r(s,\bm{a}) + \gamma \mathbb{E}[\tilde{V}^*(s^\prime)]) - \omega D_{\operatorname{KL}}(\bm{\pi}(\cdot|s) || \tilde{\bm{\pi}}^*(\cdot|s) )  
	\end{align*}
	This means that $\tilde{V}^*$ is the unique fix point of $\Gamma_\omega$. Now we have all the tools for the proof of proposition \ref{theorem:global_optimal}. And this proof is inspired by \citet{regularizer}.
	
	\begin{proof}
	    \mbox{}\par
	    Given that the initial policy is a uniform distribution, we know from Proposition \ref{converged_policy} that:
	    \begin{equation}
	        \tilde{\bm{\pi}}^*(\bm{a}|s) = \mathbbm{1}(\bm{a} \in U_s) \frac{1}{|U_s|},
	    \end{equation}
	    where $|U_s|$ is the number of actions contained in the set $U_s$. Then we consider $D_{\operatorname{KL}}(\bm{\pi}(\cdot|s) || \tilde{\bm{\pi}}^*(\cdot|s))$ and have:
	    \begin{align}
	        D_{\operatorname{KL}}(\bm{\pi}(\cdot|s) || \tilde{\bm{\pi}}^*(\cdot|s)) & = \sum_{\bm{a} \in U_s} \bm{\pi}(\bm{a}|s) \log \frac{\bm{\pi}(\bm{a}|s) }{\tilde{\bm{\pi}}^*(\bm{a}|s)} \\
	        & = \sum_{\bm{a} \in U_s} \bm{\pi}(\bm{a}|s) \log \bm{\pi}(\bm{a}|s) - \sum_{\bm{a} \in U_s} \bm{\pi}(\bm{a}|s) \log \tilde{\bm{\pi}}^*(\bm{a}|s) \\
	        & \le - \sum_{\bm{a} \in U_s} \bm{\pi}(\bm{a}|s) \log \tilde{\bm{\pi}}^*(\bm{a}|s) \quad (\bm{\pi}(\bm{a}|s) \le 1 ) \\
	        & = \log |U_s| \sum_{\bm{a} \in U_s} \bm{\pi}(\bm{a}|s) \\
	        & \le \log |U_s| \quad (\sum_{\bm{a} \in U_s} \bm{\pi}(\bm{a}|s) \le \sum_{\bm{a} \in A} \bm{\pi}(\bm{a}|s) = 1 ) \\
	        & \le \log |A| \label{KL-bound}
	    \end{align} 
	    
	    Next we will consider the relation between $\Gamma_\omega$ and $\Gamma$:
	    \begin{equation}
	        \begin{split}
	            \Gamma_\omega V(s) & = \max_{\bm{\pi}} \sum_{\bm{a}} \pi(\bm{a}|s) (r(s,\bm{a}) + \gamma \mathbb{E}[V(s^\prime)] ) - \omega D_{\operatorname{KL}}(\bm{\pi}(\cdot|s) || \tilde{\bm{\pi}}^*(\cdot|s) ) \\
	            & \le \max_{\bm{\pi}} \sum_{\bm{a}} \pi(\bm{a}|s) (r(s,\bm{a}) + \gamma \mathbb{E}[V(s^\prime)])  \\
	            & = \Gamma  V(s)
	        \end{split}
	        \label{gamma-upper}
	    \end{equation}
	    \begin{equation}
	        \begin{split}
	            \Gamma_\omega V(s) & = \max_{\bm{\pi}} \sum_{\bm{a}} \pi(\bm{a}|s) (r(s,\bm{a}) + \gamma \mathbb{E}[V(s^\prime)] ) - \omega D_{\operatorname{KL}}(\bm{\pi}(\cdot|s) || \tilde{\bm{\pi}}^*(\cdot|s) ) \\
	            & \ge \max_{\bm{\pi}} \sum_{\bm{a}} \pi(\bm{a}|s) (r(s,\bm{a}) + \gamma \mathbb{E}[V(s^\prime)]) - \max_{\bm{\pi}} D_{\operatorname{KL}}(\bm{\pi}(\cdot|s) || \tilde{\bm{\pi}}^*(\cdot|s) )   \\
	            & \ge \Gamma  V(s) - \omega \log |A| \quad (\text{From the inequality \eqref{KL-bound}} )
	        \end{split}
	        \label{gamma-lower}
	    \end{equation}
	    We will show that $\Gamma^k V(s) - \omega \log |A| \sum_{t=0}^{k - 1}\gamma^t \le \Gamma_\omega^k V(s) \le \Gamma^k V(s)$ for any $V$ and state $s$ by induction.
	
	LHS:
	\begin{align*}
	    \Gamma_\omega^{k + 1} V(s) & = \Gamma_\omega (\Gamma_\omega^{k} V(s)) \\
	    & \ge \Gamma_\omega(\Gamma^{k} V(s) - \omega \log |A| \sum_{t=0}^{k - 1}\gamma^t ) \quad (\text{From the induction and the monotonicity in lemma \ref{reg-lemma}} ) \\
	    & = \Gamma_\omega(\Gamma^{k} V(s)) -  \omega \log |A| \sum_{t=0}^{k - 1}\gamma^{t + 1}  \quad (\text{From the conclusion about constant in lemma \ref{reg-lemma}} ) \\
	    & \ge \Gamma^{k + 1} V(s) -  \omega \log |A|  -  \omega \log |A| \sum_{t=0}^{k - 1}\gamma^{t + 1}  \quad (\text{From the inequality \ref{gamma-lower} } ) \\
	    & = \Gamma^{k + 1} V(s) -\omega \log |A| \sum_{t=0}^{k}\gamma^{t}
	\end{align*}
	
	RHS:
	\begin{align*}
	    \Gamma_\omega^{k + 1} V(s) & = \Gamma_\omega (\Gamma_\omega^{k} V(s)) \\
	    & \le \Gamma_\omega(\Gamma^{k} V(s)) \quad (\text{From the induction and the monotonicity in lemma \ref{reg-lemma}} ) \\
	    & \le \Gamma^{k + 1} V(s) \quad (\text{From the inequality \ref{gamma-upper} } )
  	\end{align*}
	 Finally, with all these results above, let $k \to \infty$. As both $\Gamma_\omega$ and $\Gamma$ are $\gamma$-contraction, $\Gamma_\omega^k V \to \tilde{V}^*, \Gamma^k V \to V^*$ for any $V$. We have:
	 \begin{align*}
	     \Vert \tilde{V}^* - V^* \Vert_{\infty}  \le \omega \log |A| \sum_{t=0}^{\infty} \gamma^t 
	      = \frac{\omega}{1-\gamma} \log |A|
	 \end{align*}
	\end{proof}

	\subsection{Lemma \ref{policy_evaluation}}
	\label{sec:app-eva}
	
	\begin{proof}
		We define a new reward function 
		\begin{equation*}
			r_{\bm{\rho}}^{\bm{\pi}}(s,\bm{a}) = r(s,\bm{a}) - \omega \mathbb{E}_{s^{\prime} \sim P(\cdot |s,\bm{a} )} \left[D_{\operatorname{KL}}\left( {\bm{\pi}}(\cdot|s^{\prime}) \| {\bm{\rho}}(\cdot|s^{\prime}) \right) \right], 
		\end{equation*}
		then we can rewrite the definition of operator $\Gamma^{\operatorname{\bm{\pi}}}_{\operatorname{\bm{\rho}}}$ as 
		\begin{equation*}
			\Gamma^{\operatorname{\bm{\pi}}}_{\operatorname{\bm{\rho}}} Q(s,\bm{a})  = r_{\bm{\rho}}^{\bm{\pi}}(s,\bm{a}) + \gamma \mathbb{E}_{s^{\prime} \sim P(\cdot | s, \bm{a}), \bm{a^{\prime}} \sim  \bm{\pi}(\cdot | s^{\prime}) } \left[ Q(s^{\prime},\bm{a^{\prime}}) \right]. 
		\end{equation*}
		With this formula, we can apply the traditional convergence result of policy evaluation in \citet{RLBOOK}.
	\end{proof}
	
	\subsection{Lemma \ref{policy-theorem}}
	\label{sec:app-improv}
	
	For the proof of Lemma \ref{policy-theorem}, we need the following lemma \citep{SAC} about improving policy in entropy-regularized MDP.
	\begin{lemma}\label{improve-lemma}
		If we have a new policy ${\bm{\pi}}_{\operatorname{new}}$ and
		\[
		{\bm{\pi}}_{\operatorname{new}} = \arg \min_{\bm{\pi}} D_{\operatorname{KL}}\left( {\bm{\pi}}(\cdot|s) \|  \frac{\exp\left( Q^{{\bm{\pi}}_{\operatorname{{old}}}}_{\operatorname{ent}}(s,\cdot) / \omega \right)}{Z^{{\bm{\pi}}_{\operatorname{{old}}}}(s)} \right), 
		\] 
		where $Z^{{\bm{\pi}}_{\operatorname{{old}}}}(s)$ represents the normalization term, then we have
		\[
		Q^{{\bm{\pi}}_{\operatorname{{new}}} }_{\operatorname{ent}}(s,a) \geq Q^{{\bm{\pi}}_{\operatorname{{old}}} }_{\operatorname{ent}}(s,a),\quad \forall s \in S, a \in A.
		\]
	\end{lemma}
	
	With Lemma \ref{improve-lemma}, we have the following proof of Lemma \ref{policy-theorem}.
	
	\begin{proof}
		Let $\hat{Q}$ be the same as the definition in Proof \ref{proof1}. Then we have
		$\hat{Q}^{{\bm{\pi}}}(s,\bm{a}) = Q^{{\bm{\pi}}}_{\bm{\rho}}(s,\bm{a}) + \omega \log {\bm{\rho}}(\bm{a}|s),\,\forall {\bm{\pi}}$.
		
		According to Lemma \ref{improve-lemma},
		\begin{align*}
			&\hat{{\bm{\pi}}}_{\operatorname{{new}}}(s,\cdot) = \arg \min_{\bm{\pi}} D_{\operatorname{KL}}\left( {\bm{\pi}}(\cdot|s) \|  \frac{\exp\left( \hat{Q}^{{\bm{\pi}}_{\operatorname{{old}}}}(s,\cdot) / \omega \right)}{Z^{{\bm{\pi}}_{\operatorname{{old}}}}(s)} \right) \\
			& \hat{Q}^{\hat{{\bm{\pi}}}_{\operatorname{{new}}}}(s,\bm{a}) \geq \hat{Q}^{{\bm{\pi}}_{\operatorname{{old}}}}(s,\bm{a}),\quad \forall \bm{a} \in A.
		\end{align*}
		
		With the definition, we have 
		\begin{align*}
			& D_{\operatorname{KL}} \left( {\bm{\pi}}(\cdot|s) \|  \frac{\exp( \hat{Q}^{{\bm{\pi}}_{\operatorname{{old}}}}(s,\cdot) / \omega )}{Z^{{\bm{\pi}}_{\operatorname{{old}}}}(s)} \right)  =  D_{\operatorname{KL}} \left( {\bm{\pi}}(\cdot|s) \| {\bm{\rho}}(\cdot|s) \frac{\exp( Q^{{\bm{\pi}}_{\operatorname{{old}}}}_{\bm{\rho}}(s,\cdot) / \omega )}{Z^{{\bm{\pi}}_{\operatorname{{old}}}}(s)} \right) \\
			& {\bm{\pi}}_{\operatorname{{new}}} = \hat{{\bm{\pi}}}_{\operatorname{{new}}} \\
			& Q^{{\bm{\pi}}_{\operatorname{{new}}}}_{\bm{\rho}}(s,\bm{a}) = \hat{Q}^{{\bm{\pi}}_{\operatorname{{new}}}}(s,\bm{a}) - \omega \log {\bm{\rho}}(\bm{a}|s) \geq \hat{Q}^{{\bm{\pi}}_{\operatorname{{old}}}}(s,\bm{a}) - \omega \log {\bm{\rho}}(\bm{a}|s)  = Q^{{\bm{\pi}}_{\operatorname{old}}}_{\bm{\rho}}(s,\bm{a}) .
		\end{align*}
	\end{proof}

	\subsection{Lemma \ref{policy-LVD-theorem}}
	\label{sec:app-them-LVD}
	\begin{proof}
		From the equation 
		\begin{equation*} \label{LVD-1}
			\pi_{\operatorname{new}}^i = \arg \max_{\pi_i} \sum_{a_i} \pi_i (a_i | s) \left( k_i(s) Q_{\operatorname{\rho}}^{\pi^i_{\operatorname{old} }}(s,a_i)  - \omega \log \frac{\pi_i(a_i|s)}{\rho_i(a_i|s)}  \right),
		\end{equation*}
		we can obtain
		\begin{equation}\label{LVD-2}
			\begin{split}
				& \sum_{a_i} \pi_{\operatorname{new}}^i (a_i | s) \left( k_i(s) Q_{\operatorname{\rho}}^{\pi^i_{\operatorname{old} }}(s,a_i)  - \omega \log \frac{\pi_{\operatorname{new}}^i(a_i|s)}{\rho_i(a_i|s)}  \right) \\
				& \ge \sum_{a_i} \pi_{\operatorname{old}}^i (a_i | s) \left( k_i(s) Q_{\operatorname{\rho}}^{\pi^i_{\operatorname{old} }}(s,a_i)  - \omega \log \frac{\pi_{\operatorname{old}}^i(a_i|s)}{\rho_i(a_i|s)}  \right). \\
			\end{split}	
		\end{equation}
		By taking expectation on the both side of (\ref{LVD-2}), we can obtain the followings.
		\begin{equation} \label{LVD-3}
			\begin{split}
				& \sum_{a_{-i}} \tilde{\pi }_{-i}(a_{-i} | s) \sum_{a_i} \pi_{\operatorname{new}}^i (a_i | s) \left( k_i(s) Q_{\operatorname{\rho}}^{\pi^i_{\operatorname{old} }}(s,a_i)  - \omega \log \frac{\pi_{\operatorname{new}}^i(a_i|s)}{\rho_i(a_i|s)}  \right) \\
				& \ge \sum_{a_{-i}} \tilde{\pi }_{-i}(a_{-i} | s) \sum_{a_i} \pi_{\operatorname{old}}^i (a_i | s) \left( k_i(s) Q_{\operatorname{\rho}}^{\pi^i_{\operatorname{old} }}(s,a_i)  - \omega \log \frac{\pi_{\operatorname{old}}^i(a_i|s)}{\rho_i(a_i|s)}  \right) \\ 
				&  \forall i \in I \quad \forall \tilde{\pi }_{-i} \\
			\end{split}	
		\end{equation}
		Moreover, we can easily have the following derivation,
		\begin{equation} \label{LVD-4}
			\begin{split}
				& \sum_{a_i } \pi^{i}_{\operatorname{new}}(a_i|s)  \left( k_i(s) Q_{\operatorname{\rho}}^{\pi^i_{\operatorname{old} }}(s,a_i) + b(s)  - \omega \log \frac{\pi_{\operatorname{new}}^i(a_i|s)}{\rho_i(a_i|s)}  \right) \\
				& = \sum_{a_{-i} } u_1(a_{-i}|s) \sum_{a_i } \pi^{i}_{\operatorname{new}}(a_i|s)  \left( k_i(s) Q_{\operatorname{\rho}}^{\pi^i_{\operatorname{old} }}(s,a_i) + b(s)  - \omega \log \frac{\pi_{\operatorname{new}}^i(a_i|s)}{\rho_i(a_i|s)}  \right) \\
				& = \sum_{a_{-i} } u_2(a_{-i}|s) \sum_{a_i } \pi^{i}_{\operatorname{new}}(a_i|s)  \left( k_i(s) Q_{\operatorname{\rho}}^{\pi^i_{\operatorname{old} }}(s,a_i) + b(s)  - \omega \log \frac{\pi_{\operatorname{new}}^i(a_i|s)}{\rho_i(a_i|s)}  \right) \\
				& \forall u_1, u_2.
			\end{split}
		\end{equation}
		Then we have
		\begin{equation} \label{LVD-6}
			\begin{split}
				& \mathbb{E}_{\bm{a} \sim \bm{\pi}_{\operatorname{new}} } \left[ Q^{\bm{\pi}_{\operatorname{old}}}_{\bm{\rho}}(s,\bm{a}) -  \log \frac{{\bm{\pi}_{\operatorname{new}}}(\bm{a}|s)}{{\bm{\rho}}(\bm{a}|s)} \right]  \\ 
				& = \sum_{\bm{a}} \bm{\pi}_{\operatorname{new}}(\bm{a}|s) \sum_i \left( k_i(s) Q_{\operatorname{\rho}}^{\pi^i_{\operatorname{old} }}(s,a_i) + b(s)  - \omega \log \frac{\pi_{\operatorname{new}}^i(a_i|s)}{\rho_i(a_i|s)}  \right) \\
				& = \sum_{\bm{a}} \bm{\pi}_{\operatorname{new}}(\bm{a}|s) \sum_i \left( k_i(s) Q_{\operatorname{\rho}}^{\pi^i_{\operatorname{old} }}(s,a_i) + b(s)  - \omega \log \frac{\pi_{\operatorname{new}}^i(a_i|s)}{\rho_i(a_i|s)}  \right) \\
				& = \sum_i \sum_{a_{-i} } \pi^{-i}_{\operatorname{new}}(a_{-i}|s) \sum_{a_i } \pi^{i}_{\operatorname{new}}(a_i|s)  \left( k_i(s) Q_{\operatorname{\rho}}^{\pi^i_{\operatorname{old} }}(s,a_i) + b(s)  - \omega \log \frac{\pi_{\operatorname{new}}^i(a_i|s)}{\rho_i(a_i|s)}  \right) \\
				& = \sum_i \sum_{a_{-i} } \pi^{-i}_{\operatorname{old}}(a_{-i}|s) \sum_{a_i } \pi^{i}_{\operatorname{new}}(a_i|s)  \left( k_i(s) Q_{\operatorname{\rho}}^{\pi^i_{\operatorname{old} }}(s,a_i) + b(s)  - \omega \log \frac{\pi_{\operatorname{new}}^i(a_i|s)}{\rho_i(a_i|s)}  \right) \\
				& \ge \sum_i \sum_{a_{-i} } \pi^{-i}_{\operatorname{old}}(a_{-i}|s) \sum_{a_i } \pi^{i}_{\operatorname{old}}(a_i|s)  \left( k_i(s) Q_{\operatorname{\rho}}^{\pi^i_{\operatorname{old} }}(s,a_i) + b(s)  - \omega \log \frac{\pi_{\operatorname{old}}^i(a_i|s)}{\rho_i(a_i|s)}  \right) \\
				& = V^{\bm{\pi}_{\operatorname{old}}}_{\bm{\rho}}(s)
			\end{split}
		\end{equation}
		The fourth equation is from (\ref{LVD-4}) and the fifth inequality is from (\ref{LVD-3}).
		
		By repeatedly applying (\ref{LVD-6}) and the relation $	Q^{\bm{\pi}_{\operatorname{old}}}_{\bm{\rho}}(s,\bm{a})  = r(s,\bm{a}) + \gamma\mathbb{E}_{s^\prime} \left[ V^{\bm{\pi}_{\operatorname{old}}}_{\bm{\rho}}(s^\prime)  \right] $, we can complete the proof as followings.
		\[
		\begin{split}
			Q^{\bm{\pi}_{\operatorname{old}}}_{\bm{\rho}}(s,\bm{a}) & = r(s,\bm{a}) + \gamma \mathbb{E}_{s^\prime} \left[ V^{\bm{\pi}_{\operatorname{old}}}_{\bm{\rho}}(s^\prime)  \right] \\
			& \le r(s,\bm{a}) + \gamma\mathbb{E}_{s^\prime} \left[  \mathbb{E}_{\bm{a}^\prime \sim \bm{\pi}_{\operatorname{new}} } \left[ Q^{\bm{\pi}_{\operatorname{old}}}_{\bm{\rho}}(s^\prime,,\bm{a}^\prime) -  \log \frac{{\bm{\pi}_{\operatorname{new}}}(\bm{a}^\prime|s^\prime)}{{\bm{\rho}}(\bm{a^\prime}|s^\prime)} \right] \right] \\ 
			& = r(s,\bm{a}) + \gamma \mathbb{E}_{s^\prime} \left[  \mathbb{E}_{\bm{a}^\prime \sim \bm{\pi}_{\operatorname{new}} } \left[ r(s^\prime,\bm{a}^\prime) + \gamma \mathbb{E}_{s^{\prime\prime}} \left[ V^{\bm{\pi}_{\operatorname{old}}}_{\bm{\rho}}(s^{\prime\prime})  \right]  -  \log \frac{{\bm{\pi}_{\operatorname{new}}}(\bm{a}^\prime|s^\prime)}{{\bm{\rho}}(\bm{a}^\prime|s^\prime)} \right] \right] \\
			& \cdots \\
			& \le Q^{\bm{\pi}_{\operatorname{new}}}_{\bm{\rho}}(s,\bm{a}) \\
		\end{split}
		\]
	\end{proof}

	\subsection{Theorem \ref{policy_iteration}}
	\label{sec:app-iter}
	
	\begin{proof}
		First, we will show that Divergence Policy Iteration will monotonically improve the policy. From Lemma \ref{policy-theorem}, we know that 
		\begin{align*}
			V^{\bm{\pi}_{\operatorname{new}}}_{\bm{\rho}}(s) & = \mathbb{E}_{\bm{a} \sim \bm{\pi}_{\operatorname{new}}(\cdot | s) } \left[ 
			Q^{\bm{\pi}_{\operatorname{new}}}_{\bm{\rho}}(s,\bm{a}) - \omega \log \frac{{\bm{\pi}_{\operatorname{new}}}(\bm{a}|s)}{{\bm{\rho}}(\bm{a}|s)}
			\right] \\
			& \geq \mathbb{E}_{\bm{a} \sim \bm{\pi}_{\operatorname{new}}(\cdot | s) } \left[ Q^{\bm{\pi}_{\operatorname{old}}}_{\bm{\rho}}(s,\bm{a}) - \omega \log \frac{{\bm{\pi}_{\operatorname{new}}}(\bm{a}|s)}{{\bm{\rho}}(\bm{a}|s)} \right] \\
			& \geq \mathbb{E}_{\bm{a} \sim \bm{\pi}_{\operatorname{old}}(\cdot | s) } \left[ Q^{\bm{\pi}_{\operatorname{old}}}_{\bm{\rho}}(s,\bm{a}) - \omega \log \frac{{\bm{\pi}_{\operatorname{old}}}(\bm{a}|s)}{{\bm{\rho}}(\bm{a}|s)} \right] \\
			& = V^{\bm{\pi}_{\operatorname{old}}}_{\bm{\rho}}(s).
		\end{align*} 
		The first inequality is from the conclusion of Lemma \ref{policy-theorem} that \[Q^{\bm{\pi}_{\operatorname{new}}}_{\bm{\rho}}(s,\bm{a}) \ge Q^{\bm{\pi}_{\operatorname{old}}}_{\bm{\rho}}(s,\bm{a})\quad \forall \bm{a} \in A,\]
		and the second inequality is from the definition of $\bm{\pi}_{\operatorname{new}}$ that \[{\bm{\pi}}_{\operatorname{new}} = \arg \min_{\bm{\pi}} D_{\operatorname{KL}}\left( {\bm{\pi}}(\cdot|s) \|  \frac{\exp\left( Q^{{\bm{\pi}}_{\operatorname{{old}}}}_{\bm{\rho}}(s,\cdot) / \omega \right)}{Z^{{\bm{\pi}}_{\operatorname{{old}}}}(s)} \right).\]
		Here we have $V^{\bm{\pi}_{\operatorname{new}}}(s) \ge V^{\bm{\pi}_{\operatorname{old}}}(s),\, \forall s \in S$, and thus $J_{\bm{\rho}}(\bm{\pi}_{\operatorname{new}}) \ge J_{\bm{\rho}}(\bm{\pi}_{\operatorname{old}})$. So, Divergence Policy Iteration will monotonically improve the policy.
		
		Since the $Q^{\bm{\pi}}_{\bm{\rho}}$ is bounded above (the reward function is bounded), the sequence of Q-function $\left\{Q^{k}\right\}$ of Divergence Policy Iteration will converge and the corresponding policy sequence will also converge to some policy $\bm{\pi}_{\operatorname{conv}}$. We need to show $\bm{\pi}_{\operatorname{conv}} = \bm{\pi}^{*}_{\bm{\rho}}$.
		\begin{align*}
			V^{\bm{\pi}_{\operatorname{conv}}}_{\bm{\rho}}(s) & = \mathbb{E}_{\bm{a} \sim \bm{\pi}_{\operatorname{conv}}(\cdot | s) } \left[ 
			Q^{\bm{\pi}_{\operatorname{conv}}}_{\bm{\rho}}(s,\bm{a}) - \omega \log \frac{{\bm{\pi}_{\operatorname{conv}}}(\bm{a}|s)}{{\bm{\rho}}(\bm{a}|s)}
			\right] \\
			& \geq \mathbb{E}_{\bm{a} \sim \bm{\pi}(\cdot | s) } \left[ 
			Q^{\bm{\pi}_{\operatorname{conv}}}_{\bm{\rho}}(s,\bm{a}) - \omega \log \frac{{\bm{\pi}}(\bm{a}|s)}{{\bm{\rho}}(\bm{a}|s)}
			\right] \\
			& \geq \mathbb{E}_{\bm{a} \sim \bm{\pi}(\cdot | s) }  \bigg[ 
			r(s,\bm{a}) + \gamma \mathbb{E}_{\bm{a^{\prime}} \sim \bm{\pi}(\cdot | s^{\prime}) } \bigg[ Q^{\bm{\pi}_{\operatorname{conv}}}_{\bm{\rho}}(s^{\prime},\bm{a}^{\prime}) - \omega \log \frac{{\bm{\pi}}(\bm{a}^{\prime}|s^{\prime})}{{\bm{\rho}}(\bm{a}^{\prime}|s^{\prime})} \bigg] -\omega \log \frac{{\bm{\pi}}(\bm{a}|s)}{{\bm{\rho}}(\bm{a}|s)}
			\bigg] \\
			& \cdots \\
			& \geq V^{\bm{\pi}}_{\bm{\rho}}(s) .
		\end{align*}
		
		The first inequality is from the definition of $\bm{\pi}_{\operatorname{conv}}$ that 
		\[{\bm{\pi}}_{\operatorname{conv}} = \arg \min_{\bm{\pi}} D_{\operatorname{KL}}\left( {\bm{\pi}}(\cdot|s) \|  \frac{\exp\left( Q^{{\bm{\pi}}_{\operatorname{{conv}}}}_{\bm{\rho}}(s,\cdot) / \omega \right)}{Z^{{\bm{\pi}}_{\operatorname{{conv}}}}(s)} \right)\]
		and all the other inequalities are just iteratively using the first inequality and the relation of $Q^{\bm{\pi}}_{\bm{\rho}}$ and $V^{\bm{\pi}}_{\bm{\rho}}$. With iterations, we replace all the $\bm{\pi}_{\operatorname{conv}}$ with  $\bm{\pi}$ in the expression of $V^{\bm{\pi}_{\operatorname{conv}}}_{\bm{\rho}}(s)$ and finally we get $V^{\bm{\pi}}_{\bm{\rho}}(s)$.
		Therefore, we have
		\begin{align*}
			& V_{\bm{\rho}}^{\bm{\pi}_{\operatorname{conv}}}(s) \ge V_{\bm{\rho}}^{\bm{\pi}}(s) \quad \forall s \in S \quad \forall \bm{\pi} \in \Pi \\
			& J_{\bm{\rho}}(\bm{\pi}_{\operatorname{conv}}) \ge J_{\bm{\rho}}(\bm{\pi}) \quad \forall \bm{\pi} \in \Pi \\ 
			& \bm{\pi}_{\operatorname{conv}} = \bm{\pi}^{*}_{\bm{\rho}}.
		\end{align*}
	\end{proof}
	
	\subsection{Proposition \ref{theorem1}}
	\label{sec:app-them1}
	 We have the following proposition.
    \begin{proposition} \label{theorem1}
	If ${\bm{\pi}}^*_{\bm{\rho}} = \arg \max_{\bm{\pi}} J_{\bm{\rho}}({\bm{\pi}})$, and $V_{\bm{\rho}}^*(s) = V_{\bm{\rho}}^{{\bm{\pi}}^*_{\bm{\rho}}}(s)$ and $Q_{\bm{\rho}}^{*}(s,\bm{a}) = Q_{\bm{\rho}}^{{\bm{\pi}}^*_{\bm{\rho}}}(s,\bm{a})$ are respectively the corresponding Q-function and V-function of ${\bm{\pi}}^*_{\bm{\rho}}$, then they satisfy the following properties:
	\iffalse
	\begin{align}
		& {\bm{\pi}}^*_{\bm{\rho}}(\bm{a}|s) \propto {\bm{\rho}}(\bm{a}|s) \exp\left( \left(r(s,\bm{a}) + \gamma \mathbb{E}_{s^\prime \sim P(\cdot|s,\bm{a})}\left[ V^*_{\bm{\rho}}(s^{\prime})\right]\right) / \omega   \right) \label{opt-policy} \\
		& V^*_{\bm{\rho}}(s) = \omega \log \sum_{\bm{a}} {\bm{\rho}}(\bm{a}|s) \exp\left( \left(r(s,\bm{a}) + \gamma \mathbb{E}_{s^\prime \sim P(\cdot|s,\bm{a})}\left[ V^*_{\bm{\rho}}(s^{\prime})\right]\right) / \omega \right) \\
		& Q_{\bm{\rho}}^*(s,\bm{a})  =  r(s,\bm{a})  + \gamma \omega \mathbb{E}_{s^\prime \sim P(\cdot|s,\bm{a})}\Big[ \log \sum_{\bm{a^{\prime}}} {\bm{\rho}}(\bm{a}|s) \exp \left( Q^*_{\bm{\rho}}(s^{\prime},\bm{a}^{\prime}) / \omega \right)  \Big].
	\end{align}
	\fi
	\begin{align}
		& {\bm{\pi}}^*_{\bm{\rho}}(\bm{a}|s) \propto {\bm{\rho}}(\bm{a}|s) \exp\left( \frac{r(s,\bm{a}) + \gamma \mathbb{E}\left[ V^*_{\bm{\rho}}(s^{\prime})\right]}{\omega } \right) \label{opt-policy} \\
		& V^*_{\bm{\rho}}(s) = \omega \log \sum_{\bm{a}} {\bm{\rho}}(\bm{a}|s) \exp\left( \frac{ r(s,\bm{a}) + \gamma \mathbb{E}\left[ V^*_{\bm{\rho}}(s^{\prime})\right] }{\omega}  \right) \\
		& Q_{\bm{\rho}}^*(s,\bm{a})  =  r(s,\bm{a})  + \gamma \omega \mathbb{E}\Big[V^*_{\bm{\rho}}(s^{\prime})   \Big].
	\end{align}
    \end{proposition}
	
	Before the proof of Proposition \ref{theorem1}, we need some results about the optimal Q-function $Q^*_{\operatorname{ent}}$, the optimal V-function $V^*_{\operatorname{ent}}$, and the optimal policy ${\bm{\pi}}^*_{\operatorname{ent}}$ in entropy-regularized MDP. We have the following lemma \citep{SAC-theory}.
	\begin{lemma}\label{ent-lemma}
		\begin{align*}
			& {\bm{\pi}}^*_{\operatorname{ent}}(s,a) \propto \exp\left( \left(r(s,a) + \gamma \mathbb{E}_{s^\prime \sim P(\cdot | s, a)}\left[ V^*_{\operatorname{ent}}(s^\prime) \right]  \right) / \omega  \right) \\
			& V^*_{\operatorname{ent}}(s) = \omega \log \sum_a \exp\left( \left(r(s,a) + \gamma \mathbb{E}_{s^\prime \sim P(\cdot | s, a)}\left[ V^*_{\operatorname{ent}}(s^\prime) \right]  \right) / \omega \right) \\
			& Q^*_{\operatorname{ent}}(s,a) = r(s,a) + \gamma \omega \mathbb{E}_{s^\prime \sim P(\cdot | s, a)}\left[ \log \sum_{a^\prime} \exp\left( Q^*_{\operatorname{ent}}(s^\prime,a^\prime) / \omega \right)  \right] 
		\end{align*}
	\end{lemma}
	With Lemma \ref{ent-lemma}, we can complete the proof of Proposition \ref{theorem1}.
	
	\begin{proof}\label{proof1}
		Let $\hat{r}(s,\bm{a}) = r(s,\bm{a}) + \omega \log {\bm{\rho}}(\bm{a}|s) $, we consider the objective function
		$$
		\hat{J}({\bm{\pi}}) = \mathbb{E}_{\bm{\pi}}\left[ \sum_{t = 0} \gamma^t \left(\hat{r}(s_t,\bm{a}_t ) - \omega \log {\bm{\pi}}(\bm{a}_t|s_t) \right)  \right].$$
		Let $\hat{{\bm{\pi}}}^*(\bm{a}|s)$,$\hat{V}^*(s)$ and $\hat{Q}^*(s,\bm{a})$ be the corresponding optimal policy, V-function and Q-function of $\hat{J}({\bm{\pi}})$. By definition we can obtain
		\begin{align*}
			& \hat{{\bm{\pi}}}^*(\bm{a}|s) = {\bm{\pi}}_{\bm{\rho}}^*(\bm{a}|s) \\
			& \hat{V}^*(s) = \mathbb{E}_{\
				\bm{a} \sim \hat{{\bm{\pi}}}^*(\cdot|s) ,s^\prime \sim P(\cdot | s, a) }\left[\hat{r}(s,\bm{a}) + \gamma \hat{V}^*(s^{\prime} ) - \omega \log \hat{{\bm{\pi}}}^*(\bm{a}|s) \right] \\
			& \qquad \ \  =  \mathbb{E}_{\
				\bm{a} \sim {\bm{\pi}}_{\bm{\rho}}^*(\cdot|s), s^\prime \sim P(\cdot | s, a)}\left[r(s,\bm{a}) + \gamma \hat{V}^*(s^{\prime} ) - \omega \log \frac{{\bm{\pi}}_{\bm{\rho}}^*(\bm{a}|s )}{{\bm{\rho}}(\bm{a}|s)} \right] \\
			& \qquad \ \ = V^*_{\bm{\rho}}(s) \\
			& \hat{Q}^*(s,\bm{a}) = \hat{r}(s,\bm{a}) + \mathbb{E}_{s^\prime \sim P(\cdot | s, a)}\left[\hat{V}^*(s^{\prime}) \right] \\
			& \qquad  \quad \ \  = r(s,\bm{a})  + \mathbb{E}_{s^\prime \sim P(\cdot | s, a)}\left[V_{\bm{\rho}}^*(s^{\prime}) \right] + \omega \log {\bm{\rho}}(\bm{a} |s)   \\
			& \qquad  \quad \ \ = Q^*_{\bm{\rho}}(s,\bm{a}) + \omega \log {\bm{\rho}}(\bm{a} |s).\\	
		\end{align*}
		
		According to Lemma \ref{ent-lemma}, we have
		\begin{align*}
			& {\bm{\pi}}^*_{\bm{\rho}}(\bm{a}|s)   = \hat{{\bm{\pi}}}^*(\bm{a}|s)  \propto 
			\exp\left( \left( \hat{r}(s,\bm{a}) + \gamma \mathbb{E}_{s^\prime \sim P(\cdot | s, a)}\left[ \hat{V}^*(s^{\prime})\right]   \right) / \omega \right) \\
			& \qquad\quad\,\,\,\, =  {\bm{\rho}}(\bm{a}|s) \exp\left( \left( r(s,\bm{a}) + \gamma \mathbb{E}_{s^\prime \sim P(\cdot | s, a)}\left[ V^*_{\bm{\rho}}(s^{\prime})\right]   \right) / \omega \right)  \\
			& V^*_{\bm{\rho}}(s)   = \hat{V}^*(s) \\ 
			& \qquad\,\,\,\, = \omega \log \sum_{\bm{a}} \exp\left( \left( \hat{r}(s,\bm{a}) + \gamma \mathbb{E}_{s^\prime \sim P(\cdot | s, a)}\left[ \hat{V}^*(s^{\prime})\right]   \right) / \omega \right) \\
			& \qquad\,\,\,\,= \omega \log \sum_{\bm{a}} {\bm{\rho}}(\bm{a}|s) \exp\left( \left( r(s,\bm{a}) + \gamma \mathbb{E}_{s^\prime \sim P(\cdot | s, a)}\left[ V^*_{\bm{\rho}}(s^{\prime})\right]   \right) / \omega \right) \\
			& Q_{\bm{\rho}}^*(s,\bm{a})   = \hat{Q}^*(s,\bm{a}) - \omega \log {\bm{\rho}}(\bm{a}|s) \\
			& \qquad\quad\,\,\,\, =  \hat{r}(s,\bm{a}) + \gamma \omega \mathbb{E}_{s^\prime \sim P(\cdot | s, a)}\left[ \log \sum_{\bm{a^{\prime}}}  \exp \left( \hat{Q}^*(s^{\prime},\bm{a}^{\prime}) / \omega \right)  \right]  - \omega \log {\bm{\rho}}(\bm{a}|s) \\
			& \qquad\quad\,\,\,\, =  r(s,\bm{a}) + \gamma \omega \mathbb{E}_{s^\prime \sim P(\cdot | s, a)}\left[ \log \sum_{\bm{a^{\prime}}} {\bm{\rho}}(\bm{a}|s) \exp \left( Q^*_{\bm{\rho}}(s^{\prime},\bm{a}^{\prime}) / \omega \right)  \right] .
		\end{align*}
		
	\end{proof}
	\subsection{Proposition \ref{Q_coro}} \label{app:Q_coro}
	\begin{proof}
	    From Proposition \ref{theorem1}, we can obtain
        \iffalse
        \begin{align}
        	{\bm{\pi}}^*_{\bm{\rho}}(\bm{a}|s) & = \frac{{\bm{\rho}}(\bm{a}|s) \exp\left( \left(r(s,\bm{a}) + \gamma \mathbb{E}_{s^\prime \sim P(\cdot|s,\bm{a})}\left[ V^*_{\bm{\rho}}(s^{\prime})\right]\right) / \omega   \right)}
        	{\sum_{\bm{b}} {\bm{\rho}}(\bm{b}|s) \exp\left( \left(r(s,\bm{b}) + \gamma \mathbb{E}_{s^\prime \sim P(\cdot|s,\bm{b})}\left[ V^*_{\bm{\rho}}(s^{\prime})\right]\right) / \omega \right)}  
        	= \frac{{\bm{\rho}}(\bm{a}|s) \exp\left(  Q^*_{\bm{\rho}}(s,\bm{a}) / \omega   \right)}
        	{\exp\left(  V^*_{\bm{\rho}}(s) / \omega   \right)} \notag \\
        	& = {\bm{\rho}}(\bm{a}|s) \exp \left( \left( Q^*_{\bm{\rho}}(s,\bm{a}) - V^*_{\bm{\rho}}(s) \right) / \omega  \right). 
        \end{align}
        \fi
        \begin{align}
        	{\bm{\pi}}^*_{\bm{\rho}}(\bm{a}|s) & = \frac{{\bm{\rho}}(\bm{a}|s) \exp\left( \left(r(s,\bm{a}) + \gamma \mathbb{E}\left[ V^*_{\bm{\rho}}(s^{\prime})\right]\right) / \omega   \right)}
        	{\sum_{\bm{b}} {\bm{\rho}}(\bm{b}|s) \exp\left( \left(r(s,\bm{b}) + \gamma \mathbb{E}\left[ V^*_{\bm{\rho}}(s^{\prime})\right]\right) / \omega \right)} \notag \\
        	& = \frac{{\bm{\rho}}(\bm{a}|s) \exp\left(  Q^*_{\bm{\rho}}(s,\bm{a}) / \omega   \right)}
        	{\exp\left(  V^*_{\bm{\rho}}(s) / \omega   \right)}  \notag \\
        	& = {\bm{\rho}}(\bm{a}|s) \exp \left( \left( Q^*_{\bm{\rho}}(s,\bm{a}) - V^*_{\bm{\rho}}(s) \right) / \omega  \right). 
        \end{align}
        By rearranging the equation, we have
        \vspace{-0.2cm}
        \begin{equation}
        	V^*_{\bm{\rho}}(s) = Q^*_{\bm{\rho}}(s,\bm{a}) - \omega \log \frac{{\bm{\pi}^*_{\bm{\rho}}}(\bm{a}|s)}{{\bm{\rho}}(\bm{a}|s)},
        \end{equation}
        which is tenable for all actions $\bm{a} \in A$. 
	\end{proof}

	\subsection{Derivation of Gradient}
	\label{sec:app-grad}
	\[ 
	\begin{split} 
		\nabla_{\theta_i}\mathcal{L}_{\bm{\pi}} & = \mathbb{E}_{s \sim \mathcal{D}}\left[ \sum_{\bm{a}} \nabla_{\theta_i}{\bm{\pi}}(\bm{a}|s) \left(  Q^{{\bm{\pi}_{\operatorname{old}}}}_{\bm{\rho}}(s,\bm{a}) - \omega \log \frac{{\bm{\pi}}(\bm{a}|s)}{{\bm{\rho}}(\bm{a}|s)}  \right) + {\bm{\pi}}(\bm{a}|s)\nabla_{\theta_i}(- \omega \log \frac{{\bm{\pi}}(\bm{a}|s)}{{\bm{\rho}}(\bm{a}|s)}) \right] \\
		& = \mathbb{E}_{s \sim \mathcal{D}}\left[ \sum_{\bm{a}} {\bm{\pi}}(\bm{a}|s) \nabla_{\theta_i} \log \pi_i(a_i | s) \left(  Q^{{\bm{\pi}_{\operatorname{old}}}}_{\bm{\rho}}(s,\bm{a}) - \omega \log \frac{{\bm{\pi}}(\bm{a}|s)}{{\bm{\rho}}(\bm{a}|s)}  \right) - \omega {\bm{\pi}}(\bm{a}|s)\nabla_{\theta_i} \log \pi_i(a_i | s)  \right] \\
		& =\mathbb{E}_{s \sim \mathcal{D}, a \sim \bm{\pi}}\bigg[  \nabla_{\theta_i} \log\pi_i(a_i|s)  \bigg( Q^{{\bm{\pi}_{\operatorname{old}}}}_{\bm{\rho}}(s,\bm{a}) 
		- \omega \log \frac{{\bm{\pi}}(\bm{a}|s)}{{\bm{\rho}}(\bm{a}|s)} - \omega  \bigg) \bigg] \\
	\end{split} 
	\]
	\subsection{Theoretical Results for the Moving Average Case. }
	\label{sec:app-MA}
	    In the case where we take the target policies $\{\bm{\rho}^t\}$ as the moving average of the policies $\{\bm{\pi}^t\}$, we will formulate the iterations as followings:
    \begin{align*}
        &\boldsymbol{\rho^t} = (1-\tau)\boldsymbol{\rho^{t-1}} + \tau \boldsymbol{\pi^{t-1}} \\
        &\boldsymbol{\pi}^{t} = \arg \max_{{\boldsymbol{\pi}}} \sum_{s,\boldsymbol{a}} \mu_{\boldsymbol{\pi}}(s,\boldsymbol{a}) r(s,\boldsymbol{a})  - \omega D_{\operatorname{C}} \left( \mu_{\boldsymbol{\pi}} \| \mu_{{\boldsymbol{\rho}}^t} \right).
    \end{align*}
    Then we have:
    \begin{align*}
          J(\boldsymbol{\pi}^{t + 1})
        &\ge J(\boldsymbol{\pi}^{t + 1}) - \omega D_{\operatorname{C}} \left( \mu_{\boldsymbol{\pi}^{t+1}} \| \mu_{{\boldsymbol{\rho}}^{t+1}} \right) =  J_{{\boldsymbol{\rho}}^{t+1}}(\boldsymbol{\pi}^{t + 1}) \\ 
        & \ge J(\boldsymbol{\pi}^{t }) - \omega D_{\operatorname{C}} \left( \mu_{\boldsymbol{\pi}^{t}} \| \mu_{{\boldsymbol{\rho}}^{t+1}} \right) \\
        & \ge  J(\boldsymbol{\pi}^{t }) - (1-\tau)\omega D_{\operatorname{C}} \left( \mu_{\boldsymbol{\pi}^{t}} \| \mu_{{\boldsymbol{\rho}}^{t}} \right)
        \\
        & \ge J(\boldsymbol{\pi}^{t }) - \omega D_{\operatorname{C}} \left( \mu_{\boldsymbol{\pi}^{t}} \| \mu_{{\boldsymbol{\rho}}^{t}} \right) =  J_{{\boldsymbol{\rho}}^{t}}(\boldsymbol{\pi}^{t}).   
    \end{align*}
    The third inequality could be obtained as following:
    \begin{align*}
     & D_{\operatorname{C}} \left( \mu_{\boldsymbol{\pi}^{t}} \| \mu_{{\boldsymbol{\rho}}^{t+1}} \right)  = \sum_{s,\boldsymbol{a}} \mu_{\boldsymbol{\pi}}(s,\boldsymbol{a}) \log \frac{{\bm{\pi}}^t(\bm{a}|s)}{{\bm{\rho}}^{t+1}(\bm{a}|s)} \\
         & = \sum_{s,\boldsymbol{a}} \mu_{\boldsymbol{\pi}}(s,\boldsymbol{a}) \log \frac{{\bm{\pi}}^t(\bm{a}|s)}{\tau{\bm{\pi}}^{t}(\bm{a}|s) + (1-\tau){\bm{\rho}}^t(\bm{a}|s) } \\
         & \le \sum_{s,\boldsymbol{a}} \mu_{\boldsymbol{\pi}}(s,\boldsymbol{a}) \bigg[\log {\bm{\pi}}^t(\bm{a}|s) - \left(\tau \log {\bm{\pi}}^{t}(\bm{a}|s) + (1-\tau) \log {\bm{\rho}}^t(\bm{a}|s)  \right) \bigg] \qquad \text{from the concavity of $\log(\cdot)$} \\
         & = (1-\tau)\sum_{s,\boldsymbol{a}} \mu_{\boldsymbol{\pi}}(s,\boldsymbol{a}) \log \frac{{\bm{\pi}}^t(\bm{a}|s)}{{\bm{\rho}}^{t}(\bm{a}|s)} \\
         & = (1-\tau)D_{\operatorname{C}} \left( \mu_{\boldsymbol{\pi}^{t}} \| \mu_{{\boldsymbol{\rho}}^{t}} \right).
     \end{align*}
    So the sequence $\{J_{{\boldsymbol{\rho}}^{t}}(\boldsymbol{\pi}^{t})\}$ is still bounded and improving monotonically and will converge. With this result, we have $\tilde{Q}^{k} = Q^*_{\bm{\rho}^{k}}$ and $\tilde{V}^{k} = V^*_{\bm{\rho}^{k}}$ will converge to  $\tilde{Q}^{*}$ and $\tilde{V}^{*}$.
    
    Next we will consider the convergence of the sequence $\{\bm{\rho}^t\}$ and $\{\bm{\pi}^t\}$. From Proposition \ref{theorem1}, we have the formulation that  
    \begin{equation}
        \bm{\pi}^t(\bm{a}|s) = \frac{\bm{\rho}^{t}(\bm{a}|s) f^t(s,\bm{a})}{\sum_{\bm{b}}\bm{\rho}^{t}(\bm{b}|s) f^t(s,\bm{b})}
        \label{eq:pi-rho}
    \end{equation} 
    where $f^t(s,\bm{a})=\exp(\frac{\tilde{Q}^t(s,\bm{a})}{\omega})$ and $f^t(s,\bm{a}) \to f^*(s,\bm{a}) = \exp(\frac{\tilde{Q}^*(s,\bm{a})}{\omega})$.
    
    We define $Z^t(s) = \sum_{\bm{b}}\bm{\rho}^{t}(\bm{b}|s) f^t(s,\bm{b})$ and will show that $\{Z^t(s)\}$ will converge for each state $s$. Actually, we will show that $\{Z^t(s)\}$ monotonically improves and is bounded.
    
    With $Z^t(s)$, we could rewrite the the relation between $\bm{\rho}^{t+1}$ and $\bm{\rho}^{t}$ as followings:
    \begin{equation}
        \bm{\rho}^{t+1}(\bm{a}|s) = \left( 1 - \tau + \tau \frac{f^t(s,\bm{a})}{Z^t(s)} \right)\bm{\rho}^{t}(\bm{a}|s).
        \label{eq:rho-iteration}
    \end{equation}
    Then we have:
    \begin{align*}
       & Z^{t+1}(s) - Z^{t}(s)  = \sum_{\bm{b}}\bm{\rho}^{t+1}(\bm{b}|s) f^{t+1}(s,\bm{b}) - \sum_{\bm{b}}\bm{\rho}^{t}(\bm{b}|s) f^{t}(s,\bm{b})\\
       & \ge \sum_{\bm{b}}\bm{\rho}^{t+1}(\bm{b}|s) f^{t}(s,\bm{b}) - - \sum_{\bm{b}}\bm{\rho}^{t}(\bm{b}|s) f^{t}(s,\bm{b}) \quad \text{($f^{t+1}(s,\bm{b}) \ge f^{t}(s,\bm{b})$ from $\{\tilde{Q}^t\}$ improve monotonically.) } \\
       & = \tau \sum_{\bm{b}} \frac{ f^{t}(s,\bm{b}) - Z^t(s)}{Z^t(s)} \bm{\rho}^{t}(\bm{b}|s) f^{t}(s,\bm{b}) \qquad \text{from iteration \eqref{eq:rho-iteration}} \\
       & = \tau  \frac{ \sum_{\bm{b}} \bm{\rho}^{t}(\bm{b}|s) f^{t}(s,\bm{b})^2 - Z^t(s)^2}{Z^t(s)}  \\
       & = \tau \frac{\operatorname{Var}_{\bm{b} \sim \bm{\rho}^t(\cdot|s) } \left[ f^t(s,\bm{b}) \right] }{Z^t(s)} \\
       & \ge 0.
    \end{align*}
    Moreover, let $M^* = \max_{s,\bm{a}} f^*(s,\bm{a} ) $. When $t$ is sufficiently large we have $f^t(s,\bm{a}) \le f^*(s,\bm{a} ) + 1 \le M^* + 1$. So $Z^t(s) \le M^* + 1$ when $t$ is sufficiently large. With all the above discussions, we show that $Z^t(s)$ converge to $Z^*(s)$. 
    
    Next we will divide the actions into three classes according to the relation between $f^*(s,\bm{a})$ and $Z^*(s)$,given any fixed state $s$. Let $I^+_s = \{\bm{a} \in A| f^*(s,\bm{a}) > Z^*(s) \}$, $I^-_s = \{\bm{a} \in A| f^*(s,\bm{a}) < Z^*(s) \}$, and $I^0_s = \{\bm{a} \in A| f^*(s,\bm{a}) = Z^*(s) \}$. We will show three properties about these sets:
    \begin{itemize}
        \item[(1)] $\forall \bm{a} \in I^-_s, \quad \bm{\rho}^t(\bm{a}|s) \to 0$.
        \item[(2)] $I^+_s = \emptyset$.
        \item[(3)] $\sum_{\bm{a} \in I^0_s} \rho^t(\bm{a}|s) \to 1$.
    \end{itemize}
    It is obvious that the property (3) is an corollary of property (1) and (2). So we will focus on property (1) and (2).
    
    As for property (1), consider any action $\bm{a} \in I^-_s$. Let $\epsilon = \frac{Z^*(s) - f^*(s,\bm{a})}{4} > 0$, when $t$ is sufficiently large, we have $f^t(s,\bm{a}) < f^*(s,\bm{a}) + \epsilon = Z^*(s) - 3\epsilon$ and $Z^t(s) > Z^*(s) - \epsilon $. Then we have 
    \begin{align*}
        \bm{\rho}^{t+1}(\bm{a}|s) & = \left( 1 - \tau + \tau \frac{f^t(s,\bm{a})}{Z^t(s)} \right)\bm{\rho}^{t}(\bm{a}|s) \\
        &< \left( 1 - \tau + \tau \frac{Z^*(s) - 3\epsilon}{Z^*(s) + \epsilon} \right)\bm{\rho}^{t}(\bm{a}|s) \\
        & = \left( 1 - \frac{ 4\epsilon\tau}{Z^*(s) + \epsilon} \right)\bm{\rho}^{t}(\bm{a}|s).
    \end{align*}
    Since the constant $ 1 - \frac{ 4\epsilon\tau}{Z^*(s) + \epsilon} < 1$, we know that $\bm{\rho}^t(\bm{a}|s) \to 0$ for action $\bm{a}$.
    
    As for property (2), suppose that $I^+_s \not = \emptyset$. Then, we could take some action $\bm{a} \in I^+_s$. Let $\epsilon = \frac{f^*(s,\bm{a}) - Z^*(s)}{4} > 0$, when $t$ is sufficiently large, we have $f^t(s,\bm{a}) > f^*(s,\bm{a}) - \epsilon$ and $Z^t(s) < Z^*(s) + \epsilon = f^*(s,\bm{a}) - 3\epsilon $. Then we have
    \begin{align*}
        \bm{\rho}^{t+1}(\bm{a}|s) & = \left( 1 - \tau + \tau \frac{f^t(s,\bm{a})}{Z^t(s)} \right)\bm{\rho}^{t}(\bm{a}|s) \\
        &> \left( 1 - \tau + \tau \frac{f^*(s,\bm{a}) - \epsilon}{f^*(s,\bm{a}) - 3\epsilon} \right)\bm{\rho}^{t}(\bm{a}|s) \\
        & = \left( 1 + \frac{ 2\epsilon\tau}{f^*(s,\bm{a}) - 3\epsilon} \right)\bm{\rho}^{t}(\bm{a}|s).
    \end{align*}
     Since the constant $ 1 + \frac{ 2\epsilon\tau}{f^*(s,\bm{a}) - 3\epsilon} > 1$, we know that $\bm{\rho}^t(\bm{a}|s) \to \infty$ for action $\bm{a}$. But we know that $\bm{\rho}^t(\bm{a}|s) \le 1$ which is a contradiction. 
     
     From these three properties we actually know that $Z^*(s) = \max_{\bm{a}}f^*(s,\bm{a})$. The proof is as followings: from property (2) we know that $f^*(s,\bm{a}) \le Z^*(s), \forall \bm{a} \in A$; suppose that  $f^*(s,\bm{a}) < Z^*(s), \forall \bm{a} \in A$, then from property (1) we know that $\sum_{\bm{a}} \bm{\rho}^t(\bm{a}|s) \to 0$ which contradicts to $\sum_{\bm{a}} \bm{\rho}^t(\bm{a}|s)  = 1$. This fact could also be written as $I^0_s = U_s$, where $U_s$ is the same definition as the proof for Proposition \ref{converged_policy} in Appendix~\ref{app:converged_policy}
     
     Finally, with all these preparations above, we could discuss the convergence of $\bm{\pi}^t$. 
     
     From iteration \eqref{eq:rho-iteration}, we could obtain the following formula:
     \begin{equation}
        \bm{\rho}^t(\bm{a}|s) = \bm{\rho}^0(\bm{a}|s)f^t(s,\bm{a}) \Pi_{k=0}^t \left(1-\tau + \tau \frac{f^k(s,\bm{a})}{Z^k(s)} \right) = \bm{\pi}^0(\bm{a}|s)f^t(s,\bm{a}) \Pi_{k=0}^t \left( 1-\tau + \tau \frac{f^k(s,\bm{a})}{Z^k(s)} \right). 
     \end{equation}
    Combining this with \eqref{eq:pi-rho}, we have
    \begin{align*}
        \bm{\pi}^t(\bm{a}|s) & = \frac{\bm{\pi}^0(\bm{a}|s)f^t(s,\bm{a}) \Pi_{k=0}^{t-1} \left( 1-\tau + \tau \frac{f^k(s,\bm{a})}{Z^k(s)} \right)}{\sum_{\bm{b}}\bm{\pi}^0(\bm{b}|s)f^t(s,\bm{b}) \Pi_{k=0}^{t-1} \left( 1-\tau + \tau \frac{f^k(s,\bm{b})}{Z^k(s)} \right)} \\
        & = \frac{\bm{\pi}^0(\bm{a}|s)f^t(s,\bm{a}) \Pi_{k=0}^{t-1} \left( (1-\tau) Z^k(s) + \tau f^k(s,\bm{a})\right)}{\sum_{\bm{b}}\bm{\pi}^0(\bm{b}|s)f^t(s,\bm{b}) \Pi_{k=0}^{t-1} \left( (1-\tau) Z^k(s) + \tau f^k(s,\bm{b}) \right)} \\
        & = \frac{\bm{\pi}^0(\bm{a}|s) \frac{f^t(s,\bm{a})\Pi_{k=0}^{t-1} \left( (1-\tau) Z^k(s) + \tau f^k(s,\bm{a})\right)}{\left( Z^*(s) \right)^{t+1}} }{\sum_{\bm{b}}\bm{\pi}^0(\bm{b}|s)\frac{f^t(s,\bm{b})\Pi_{k=0}^{t-1} \left( (1-\tau) Z^k(s) + \tau f^k(s,\bm{b})\right)}{\left( Z^*(s) \right)^{t+1}}}.
    \end{align*}
    We also have 
    \begin{equation}
        \lim_{t \to \infty}\left( (1-\tau)Z^t(s) + \tau f^t(s,\bm{a}) \right) = \begin{cases}
           Z^*(s) & \bm{a} \in I^0_s \\
           (1-\tau)Z^*(s) + \tau f^*(s,\bm{a}) < Z^*(s) & \bm{a} \in I^-_s
        \end{cases}.
    \end{equation}
    Similarly to the proof for Proposition \ref{converged_policy}, we have
    \begin{equation}
        \lim_{t \to \infty}\frac{f^t(s,\bm{a})\Pi_{k=0}^{t-1} \left( (1-\tau) Z^k(s) + \tau f^k(s,\bm{a})\right)}{\left( Z^*(s) \right)^{t+1}} = \begin{cases}
            1 & \bm{a} \in I^0_s \\
            0 & \bm{a} \in I^-_s
        \end{cases}.
    \end{equation}
    Finally we obtain 
    \begin{equation}
        \tilde{\bm{\pi}}^*(\bm{a}|s) = \lim_{t \to \infty} \bm{\pi}^t(\bm{a}|s) = \mathbbm{1}{(\bm{a} \in I^0_s)} \frac{\bm{\pi}^0(\bm{a}|s)}{\sum_{\bm{b} \in I^0_s}\bm{\pi}^0(\bm{b}|s)} = \mathbbm{1}{(\bm{a} \in U_s)} \frac{\bm{\pi}^0(\bm{a}|s)}{\sum_{\bm{b} \in U_s}\bm{\pi}^0(\bm{b}|s)}.
    \end{equation}
    This conclusion is actually the same as Proposition \ref{converged_policy}.
    
    Moreover, as $\bm{\rho}^t$ is the moving average of $\bm{\pi}^t$, it is obvious that $\bm{\rho}^t \to \tilde{\bm{\pi}}^*$. With this result, we could define the same operator $\Gamma_\omega$ as the proof for Theorem \ref{theorem:global_optimal} in Appendix~\ref{app:global_optimal} and use the same proof to show that $\tilde{V}^*$ is the fixed point of $\Gamma_\omega$ and finally obtain the same conclusion of Theorem \ref{theorem:global_optimal}.
\section{Algorithm}
\label{app:algo}
	
	Algorithm \ref{alg1} gives the training procedure of DMAC.
	
	\begin{algorithm}[h]
		\caption{DMAC}
		\label{alg1}
		\begin{algorithmic}[1]
			\FOR{episode = $1$ to $m$}
			\STATE Initialize the environment and receive initial state $s$ \
			\FOR{$t = 1$ to max-episode-length}
			\STATE For each agent $i$, select action $a_i \sim \pi_i(\cdot|s)$ \
			\STATE Execute joint-action $\bm{a} = (a_1,a_2,\cdots,a_n)$ and observe reward $r$ and next state $s^{\prime}$\
			\STATE Store $(s,\bm{a},r,s^{\prime})$ in replay buffer $\mathcal{D}$ \
			\ENDFOR
			\STATE Sample a random minibatch of $K$ samples from $\mathcal{D}$, $\{(s_k,\bm{a}_k,r_k,s^{\prime}_k)\}_K$ \
			\FOR{agent $i = 1$ to $n$}	
			\STATE Update policy $\pi_i$: $\theta_i = \theta_i + \beta \nabla_{\theta_i}
			\mathcal{L}_{\bm{\pi}}$	\ 	
			\STATE Update target policy $\rho_i$: $\tilde{\theta}_i= (1-\tau)\tilde{\theta}_i + \tau \theta_i$ \
			\ENDFOR
			\STATE Update critic: $\phi = \phi - \alpha \nabla_\phi \mathcal{L}_Q$\
			\STATE Update target critic: $\tilde{\phi} =(1-\tau) \tilde{\phi}+\tau \phi$ \
			\ENDFOR
		\end{algorithmic}
	\end{algorithm}

\section{Implementation Details}
	\label{app:details}
	SMAC is an MARL environment based on the game StarCraft II (SC2). Agents control different units in SC2 and can attack, move or take other actions. The general mode of SMAC tasks is that agents control a team of units to counter another team controlled by built-in AI. The target of agents is to wipe out all the enemy units and agents will gain rewards when hurting or killing enemy units. Agents have an observation range and can only observe information of units in this range, but the information of all the units can be accessed in training. We test all the methods in totally 8 tasks/maps: 3m, 2s3z, 3s5z, 8m, 1c3s5z, 3s\_vs\_3z, 2c\_vs\_64zg, and MMM2.

\subsection{Modifications of the Baseline Methods}
\label{app:modifications}
	The modifications of the baseline methods, COMA, MAAC, QMIX, DOP, and FOP, are as follows:
	\begin{itemize}
		\item \textsf{COMA.} We keep the original critic and actor networks and add a target policy network with the same architecture as the actor. As COMA is on-policy but COMA+DMAC is off-policy, we add a replay buffer for experience replay.
		\item \textsf{MAAC} already has a target policy for stability, so we do not need to modify the network architecture. We only change the update rule for the critic and actors.
		\item \textsf{QMIX} is a value-based method, so we need to add a policy network and a target policy network for each agent. We keep the original individual Q-functions to learn the critic in QMIX+DMAC. In divergence-regularized MDP, the $\max$ operator is not needed in the critic update, so we abandon the hypernet and use an MLP, which takes individual Q-values and state as input  and produces the joint Q-value. This architecture is simple and its expressive capability is not limited by QMIX's IGM condition.
		\item \textsf{DOP.} We keep the original critic and actor networks and add a target policy network with the same architecture as the actor. We keep the value decomposition structure and use off-policy TD($\lambda$) for all samples in training to replace the tree backup loss and on-policy TD($\lambda$) loss.
		\item \textsf{FOP.} We replace the entropy regularizers with divergence regularizers in FOP and use the update rules of DMAC. We keep the original architecture of FOP.
		%\vspace{-0.10cm}
	\end{itemize}

\subsection{Hyperparameters}
\label{app:hyper}
	As all tasks in our experiments are cooperative with shared reward, so we use parameter-sharing policy network and critic network for MAAC and MAAC+DMAC to accelerate training. Besides, we add a RNN layer to the policy network and critic network in MAAC and MAAC+DMAC to settle the partial observability. 
	
	All the policy networks are the same as two linear layers and one GRUCell layer with ReLU activation and the number of hidden units is 64. The individual Q-networks for QMIX group is the same as the policy network mentioned before. The critic network for COMA group is a MLP with three 128-unit hidden layers and ReLU activation. The attention dimension in the critic networks of MAAC group is 32. The number of hidden units of mixer network in QMIX group is 32. The learning rate for critic is $10^{-3}$ and the learning rate for actor is $10^{-4}$.  We train all networks with RMSprop optimizer. The discouted factor is $\gamma = 0.99$. The coefficient of regularizer is $\omega = 0.01$ for SMAC tasks and $\omega = 0.2$ for the stochastic game. The \textit{td\_lambda} factor used in COMA group is $0.8$. The parameter used for soft updating target policy is $\tau = 0.01$. Our code is based on the implementation of PyMARL \citep{SMAC}, MAAC \citep{MAAC}, DOP \citep{DOP}, FOP \citep{FOP}  and an open source code for algorithms in SMAC (\url{https://github.com/starry-sky6688/StarCraft}).
	
	\begin{figure*}[t]
		\centering
		\includegraphics[width = 1\textwidth]{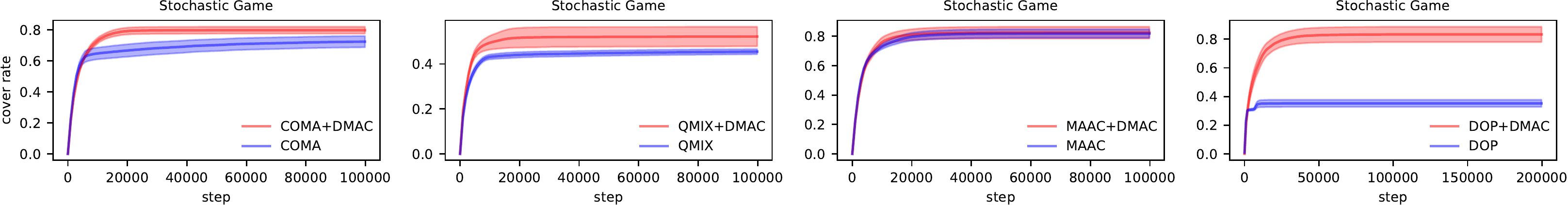}
		\vspace*{-0.8cm} 
		\caption{Learning curves in terms of cover rates of COMA, MAAC, QMIX and DOP groups in the randomly generated stochastic game.}
		\label{four-cover-rate} 
	    %\vspace*{-0.2cm} 
	\end{figure*}
	
	\begin{figure*}[!t]
		\centering
		\includegraphics[width = 0.6\textwidth]{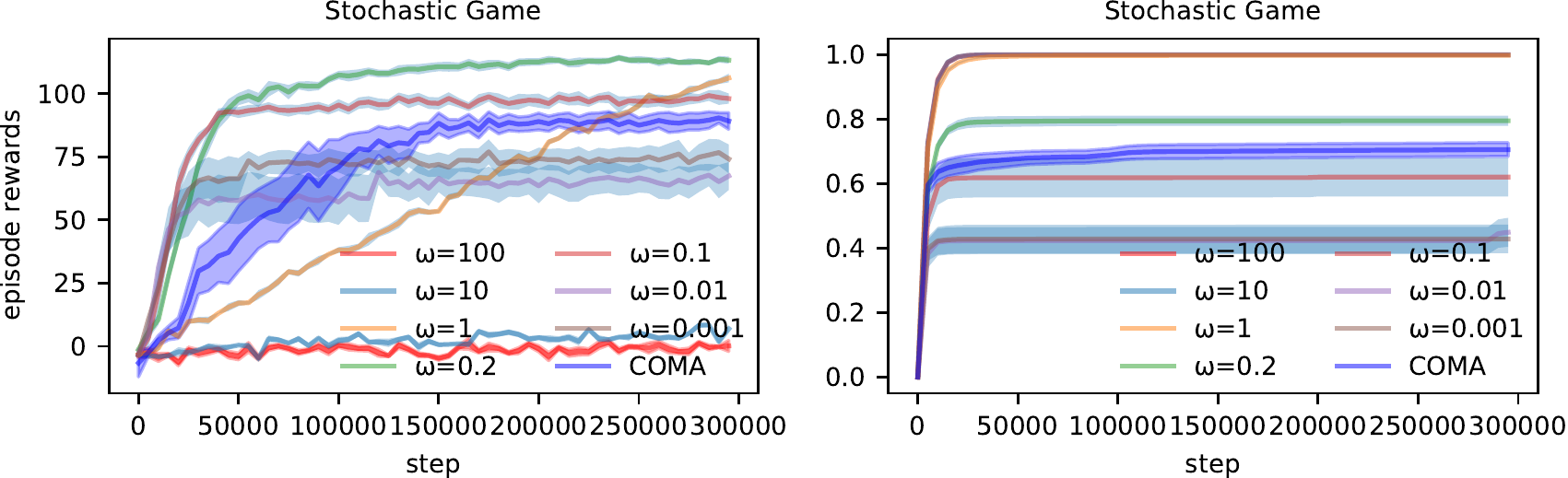}
		\vspace*{-0.4cm}
		\caption{Learning curves in terms of cover rates of COMA and COMA+DMAC with different $\omega$ in the randomly generated stochastic game.}
		\label{learning-curve-omega}
		\vspace*{-0.2cm} 
	\end{figure*}

\subsection{Experiment Settings}
	We trained each algorithms for five runs with different random seeds. In SMAC tasks, we train each algorithm for one million steps in each run for COMA, QMIX, MAAC, and DOP groups (except 2c\_vs\_64zg and MMM2) and two million steps for FOP groups. We evaluate 20 episodes in every 10000 training steps in the one million steps training procedure and in every 20000 steps in the two million steps training procedure. In evaluation, we select greedy actions for QMIX and FOP (following the setting in the FOP paper) and sample actions according to action distribution for stochastic policy (COMA, MAAC, DOP and divergence-regularized methods). We do all the experiments by a server with 2 NVIDIA A100 GPUs.

	\section{Additional Results}
	\label{app:add}
	
	%Figure~\ref{five-matrix} shows the learning curves of all five groups in terms of mean episode rewards and cover rates in the randomly generated stochastic game. 

	Figure~\ref{four-cover-rate} shows the learning curves in terms of cover rates of COMA, QMIX, MAAC and DOP groups in the randomly generated stochastic game. 
	%We need to argue that the exploration ability of QMIX relies on the $\epsilon$-greedy method and is affected by the choice of the hyperparameter $\epsilon$. If we change related hyperparameters such as the initial value of $\epsilon$ and the decay speed of $\epsilon$, we will get a different cover rate result. So we do not compare the cover rate results of QMIX and QMIX+DMAC and this result is for reference only .
		
	Figure~\ref{learning-curve-omega} shows learning curves in terms of cover rates and episode rewards of COMA and COMA+DMAC with different $\omega$ in the randomly generated stochastic game.
	
\begin{figure*}[t]
		\centering
		\includegraphics[width = 1\textwidth]{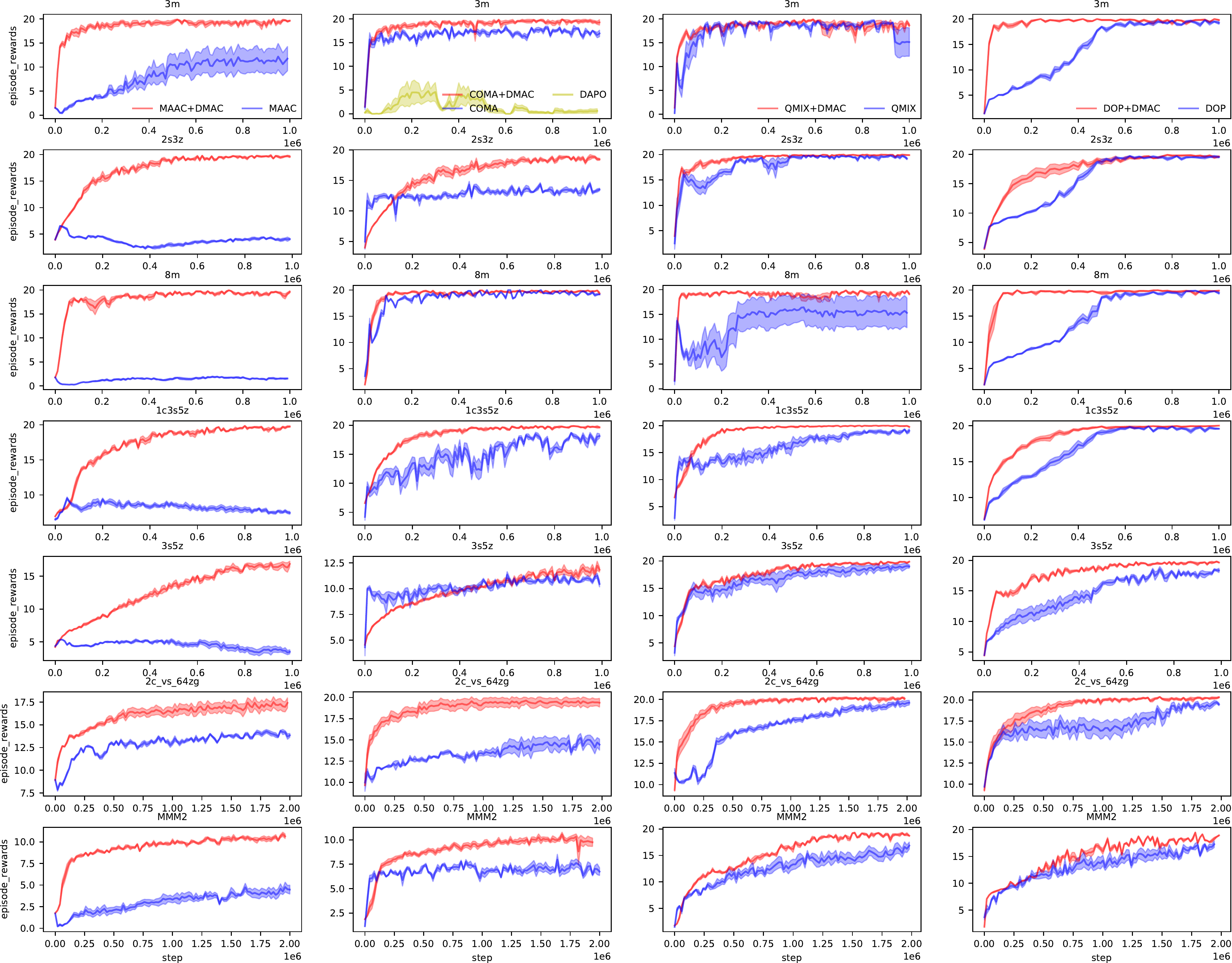}
	    \vspace*{-0.6cm}
		\caption{Learning curves in terms of mean episode reward of COMA, MAAC, QMIX, and DOP groups in seven SMAC maps (each row corresponds to a map and each column corresponds to a group). }
		\label{SMAC--reward-curve}
		\vspace*{-0.2cm} 
	\end{figure*}

\begin{figure*}[t]
		\setlength{\abovecaptionskip}{2pt}
		\centering
		\includegraphics[width = 1\textwidth]{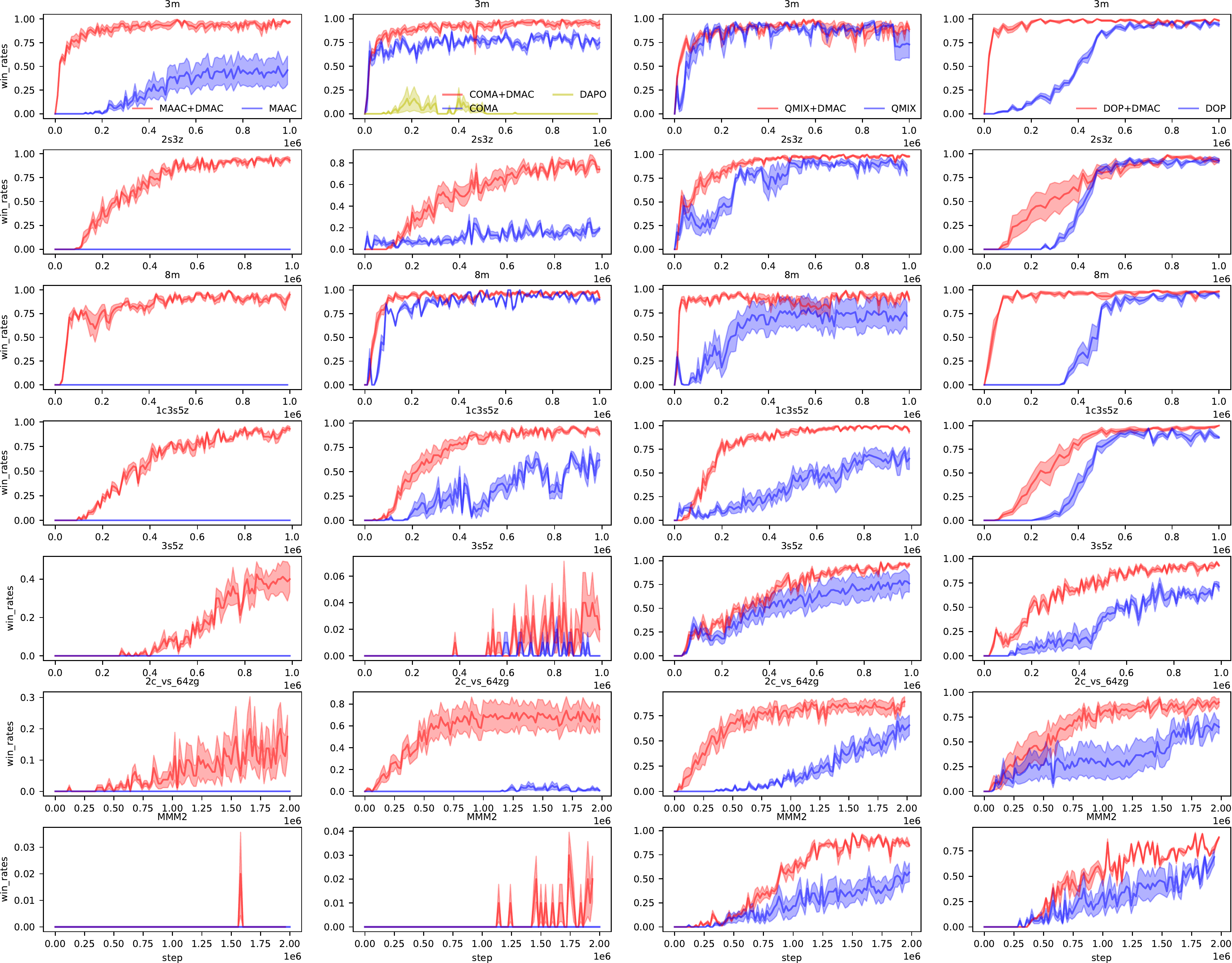}
		\vspace*{-0.6cm}
		\caption{Learning curves in terms of win rate of COMA, MAAC, QMIX, and DOP groups in seven SMAC maps (each row corresponds to a map and each column corresponds to a group). }
		\label{more-SMAC-curve}
		%\vspace*{-0.2cm} 
	\end{figure*}

	Figure~\ref{SMAC--reward-curve} and Figure~\ref{more-SMAC-curve} shows the learning curves of COMA, MAAC, QMIX and DOP groups in terms of mean episode rewards and win rates in seven SMAC maps.

\begin{figure*}[t]
		\setlength{\abovecaptionskip}{2pt}
		\centering
		\includegraphics[width = 1\textwidth]{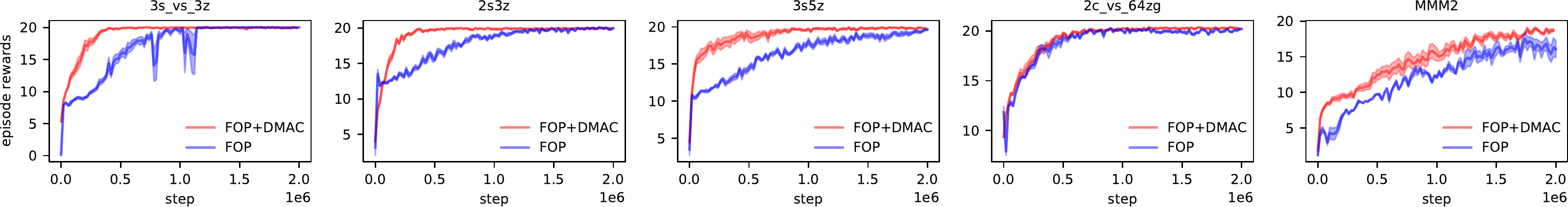}
		\vspace*{-0.8cm}
		\caption{Learning curves in terms of mean episode rewards of FOP+DMAC and FOP in five SMAC maps (each column corresponds to a map).}
		\label{fop-smac-episode-rewards}
		\vspace*{-0.2cm} 
	\end{figure*}

	Figure~\ref{fop-smac-episode-rewards} shows the learning curves of FOP+DMAC and FOP in terms of mean episode rewards in 3s\_vs\_3z,2s3z,3s5z, 2c\_vs\_64zg and MMM2. 
	
\begin{figure*}[t]
		\centering
		\includegraphics[width = 0.6\textwidth]{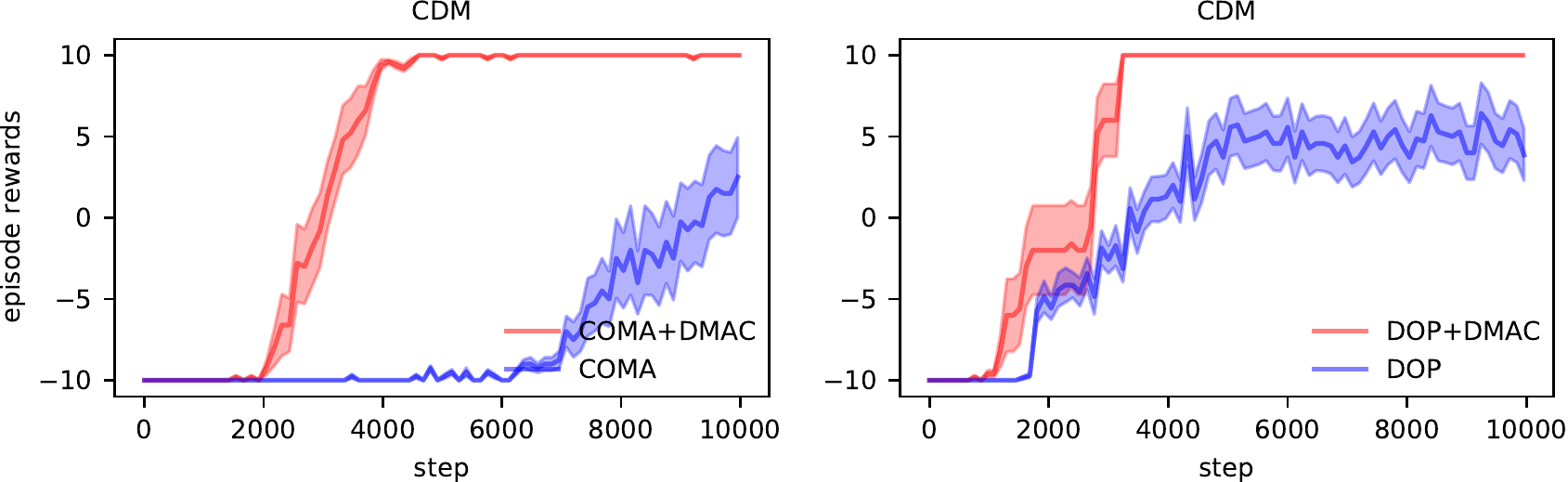}
		\vspace*{-0.4cm}
		\caption{Learning curves in terms of mean episode rewards of COMA and DOP groups in the CDM environment used by the DOP paper.}
		\label{CDM-episode-rewards}
		\vspace*{-0.2cm} 
	\end{figure*}

	Figure~\ref{CDM-episode-rewards} shows the learning curves in terms of mean episode rewards of COMA and DOP groups in the CDM environment used by the DOP paper.

\end{document}